\documentclass[pdflatex,sn-mathphys-num]{sn-jnl}

\usepackage{graphicx}%
\usepackage{multirow}%
\usepackage{amsmath,amssymb,amsfonts}%
\usepackage{amsthm}%
\usepackage{mathrsfs}%
\usepackage[title]{appendix}%
\usepackage{xcolor}%
\usepackage{textcomp}%
\usepackage{manyfoot}%
\usepackage{booktabs}%
\usepackage{algorithm}%
\usepackage{algorithmicx}%
\usepackage{algpseudocode}%
\usepackage{listings}%
\usepackage{geometry}
\usepackage{graphicx}
\usepackage{subcaption}
\theoremstyle{thmstyleone}%
\theoremstyle{thmstyletwo}%

\theoremstyle{thmstylethree}%

\begin{document}

\title[Article Title]{OceanLight: Efficient Global Ocean Forecasting via Geometry-Adaptive Unstructured Mesh Representation}


\author[1]{\fnm{Wei} \sur{Wu}}

\author*[2]{\fnm{Xiang} \sur{Wang}}\email{xiangwangcn@nudt.edu.cn}

\author[2]{\fnm{Hongze} \sur{Leng}}

\author[3]{\fnm{Qingye} \sur{Min}}

\author[2]{\fnm{Junxing} \sur{Zhu}}

\author*[12]{\fnm{Junqiang} \sur{Song}}\email{junqiang@nudt.edu.cn}

\affil[1]{\orgdiv{College of Computer Science and Technology}, \orgname{National University of Defense Technology}, \orgaddress{\street{Deya Street}, \city{Changsha}, \postcode{410073}, \state{Hunan}, \country{China}}}

\affil[2]{\orgdiv{College of
Meteorology and Oceanography}, \orgname{National University of Defense Technology}, \orgaddress{\street{Deya Street}, \city{Changsha}, \postcode{410073}, \state{Hunan}, \country{China}}}

\affil[3]{\orgname{Hunan Institute of Advanced Technology}, \orgaddress{\street{Qingshan Street}, \city{Changsha}, \postcode{410205}, \state{Hunan}, \country{China}}}


\abstract{Reliable global ocean forecasting is critical for climate monitoring, marine navigation, and extreme event early warning. Physics-based ocean forecasting models impose prohibitive computational costs, while existing deep learning approaches predominantly rely on structured-grid architectures, incurring unnecessary computation on masked land cells and enforcing uniform resolution across dynamically heterogeneous ocean regions regardless of local flow complexity. Here we present OceanLight, an efficient global ocean forecasting framework innovatively combining geometry-adaptive unstructured mesh tokenization with a graph neural network (GNN) backbone. OceanLight achieves pointwise forecast accuracy and kinetic energy spectral fidelity exceeding both operational numerical analyses and state-of-the-art AI-based models, while surpassing all AI-based ocean models in geostrophic balance consistency. Furthermore, OceanLight demonstrates reliable mesoscale eddy representation, capturing coherent ocean structures beyond pointwise statistical optimization. These capabilities are delivered with a 62\% reduction in GPU memory consumption and 70\% reduction in FLOPs relative to structured-grid baselines. Our unstructured mesh representation establishes a generalizable paradigm for scalable data-driven oceanography.}

\maketitle

\section{Introduction}\label{sec1}

The ocean plays a fundamental role in Earth’s climate system by storing and transporting heat, freshwater, and momentum~\cite{heat, oceanwarming}, thereby influencing weather patterns, sea level variability, and marine ecosystems~\cite{transport, cmemsreview}. Reliable ocean forecasting is essential for a broad range of scientific and societal applications, including climate monitoring and prediction, marine navigation and fisheries management~\cite{managing}, as well as early warning for extreme events such as storm surges and coastal flooding\citep{warnings, le2019observation}. In addition, ocean forecasts provide critical boundary and initial conditions for coupled Earth system models, and their accuracy directly affects the performance of weather and climate predictions\cite{tonani2015status}.

State-of-the-art operational ocean forecasting relies on physics-based general circulation models (GCMs) such as NEMO~\cite{NEMO}, MOM6~\cite{MOM6}, and HYCOM~\cite{HYCOM, HYCOMsimulation}, which discretize the governing primitive equations onto structured latitude–longitude or tripolar grids. While these models encode well-established oceanic dynamics, their computational demands are prohibitive: global eddy-resolving simulations at high resolution require large-scale HPC resources\cite{smith2000pop,hurlburt2008eddyresolving,hewitt2017highres} and impose substantial wall-clock and throughput costs\cite{chassignet2020omip2,chassignet2021importance,seo2023ocean}, limiting ensemble size, reanalysis throughput and real-time applicability~\cite{jean2021copernicus, lellouche2013evaluation}. 

The success of deep learning in atmospheric forecasting has stimulated growing interest in data-driven alternatives for ocean prediction. Models such as FourCastNet~\cite{fourcastnet}, Pangu-Weather~\cite{Pangu}, and GraphCast~\cite{graphcast} have demonstrated that neural networks trained on reanalysis data can produce skillful medium-range atmospheric forecasts at a fraction of the cost of numerical weather prediction. Inspired by these results, recent efforts have begun applying deep learning architectures to ocean state prediction~\cite{xihe, wenhai, glonet, langya, oceanforecastbench}. However, directly transferring the atmospheric paradigm to the global ocean domain introduces several structural challenges that are absent or less severe in the atmosphere. The ocean domain is geometrically complex, multiply-connected, and characterized by sharp bathymetric gradients~\cite{FESOM}; applying structured-grid neural architectures to irregular ocean boundaries may introduce systematic discretization errors and require costly preprocessing (masking, interpolation, or padding) which risks distorting the learned spatial relationships, particularly near coastlines and islands~\cite{challenge, CNNIssues, VoronoiInduced}. Furthermore, uniform structured tokenization allocates equal computational resources to dynamically quiescent regions (the abyssal plain) and energetically active regions (boundary currents, eddy fields), leading to unnecessary memory and compute overhead~\cite{multiresolution, variableresolution}. Notably, few prior works have proposed a data-driven global ocean forecasting framework based on an unstructured spatial representation, which can naturally conform both to ocean geometry and to spatial heterogeneity~\cite{gnnoceanforecasting, GNNMediterraneanforecasting, gnn-surrogate}.

To address the aforementioned limitations of structured-grid ocean forecasting paradigms, we develop OceanLight, a novel global ocean forecasting framework that integrates principled unstructured mesh tokenization with a graph neural network (GNN) backbone~\cite{gnn1, gnn2}. Taking comprehensive 0.25° ocean state variables—including sea surface height, as well as temperature, salinity, and current velocities across 23 vertical depth levels—as inputs, OceanLight produces deterministic forecasts for lead times of 1–10 days, with individual models optimized for each forecasting horizon. The framework of OceanLight is demonstrated in figure 1.

The key innovations of this work are threefold. First, we design OceanLight as a novel climatology-aware unstructured mesh forecasting framework for global ocean prediction. Quantitative results confirm its superiority over traditional numerical models and the state-of-the-art GLONET baseline~\cite{glonet}. OceanLight matches numerical simulations in geostrophic balance consistency, leads all competitors in kinetic energy spectral fidelity, and accurately reconstructs coherent mesoscale eddies. Second, we introduce an unstructured mesh representation derived from global ocean climatology. This representation naturally adapts to complex ocean boundaries and dynamic heterogeneity, refining resolution in dynamically active regions and coarsening resolution in tranquil basins automatically. Third, by equipping the GNN framework with unstructured mesh representation, OceanLight achieves a 73\% reduction in FLOPs and a 62\% drop in GPU memory consumption relative to the original structured-grid GNN-based GraphCast, thus facilitating efficient training for global ocean forecasting on a single consumer-grade GPU.
\begin{figure}[H]
    \centering
    \includegraphics[width=\textwidth]{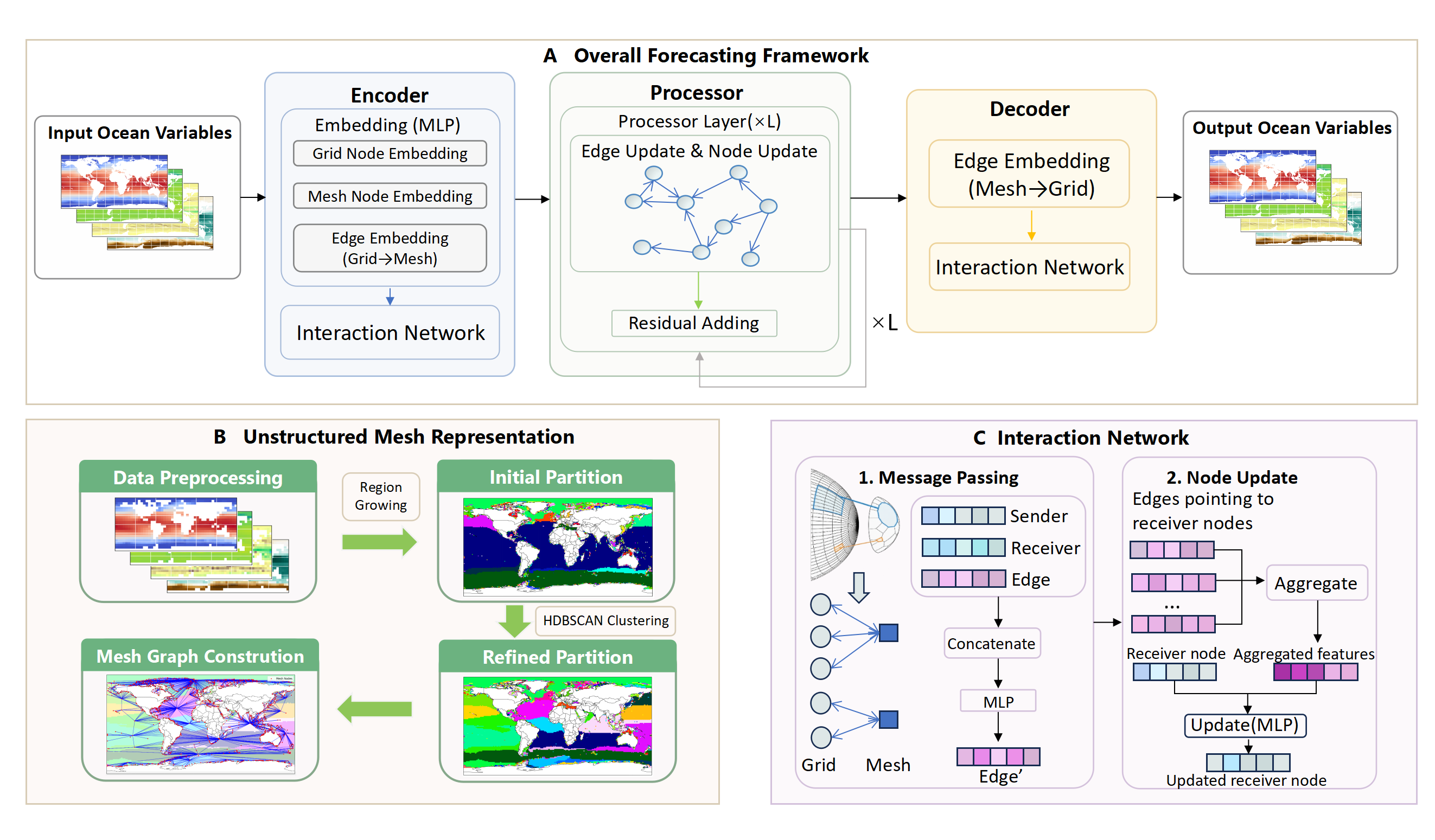}
    \caption{Overall architecture of OceanLight. OceanLight couples a geometry-adaptive unstructured mesh representation with a graph neural network (GNN) built on the encode–process–decode paradigm to forecast global ocean states. Input ocean variables at 0.25° resolution are encoded onto a compact unstructured mesh, propagated through an L-layer message-passing processor, and decoded back onto the original grid to produce the output ocean variables.\textbf{(A)} Overall forecasting framework. The encoder embeds grid nodes, mesh nodes, and grid-to-mesh edges via separate MLPs and passes information from grid to mesh through an Interaction Network. The processor applies L unshared message-passing layers, each performing edge and node updates followed by residual connection and LayerNorm on the mesh representation. The decoder embeds mesh-to-grid edges via an MLP and uses an Interaction Network to propagate the processed mesh representation back to the grid nodes, yielding the output ocean variables.\textbf{(B)} Unstructured mesh representation. Starting from ocean climatology data preprocessing, an initial partition of the global ocean is obtained via a region-growing algorithm guided by gradient-based seed selection. Large regions area are further refined through HDBSCAN clustering to produce the final refined partition. Each resulting region is treated as a mesh node, and mesh graph construction connects nodes based on geographical adjacency and multi-step (k-hop) connectivity, yielding the final mesh graph used by the GNN backbone.\textbf{(C)} Interaction Network. For each directed edge connecting a sender and a receiver node (e.g., from grid to mesh), 1. Message Passing concatenates the sender, edge, and receiver features and passes them through an MLP to produce an updated edge feature ($\mathrm{Edge}^{\prime}$). 2. Node Update aggregates the updated edge features incoming to each receiver node (by summation) and combines the aggregated features with the original receiver node feature through a second MLP to produce the updated receiver node representation.}
    \label{fig:framework}
\end{figure}

\section{Results}\label{sec2}

\subsection{Forecast accuracy across variables and lead times}
We compare our model with GLONET and graphcast, using oceanbench~\citep{oceanbench} as inputs and make predictions across different lead times. We calculate RMSE and ACC to evaluate basic forecast accuracy across variables and lead times. The result of RMSE and ACC of different variables is shown in figure 2.

\begin{figure}[H]
    \centering
    \begin{subfigure}{\textwidth}
        \centering
        \includegraphics[width=\textwidth]{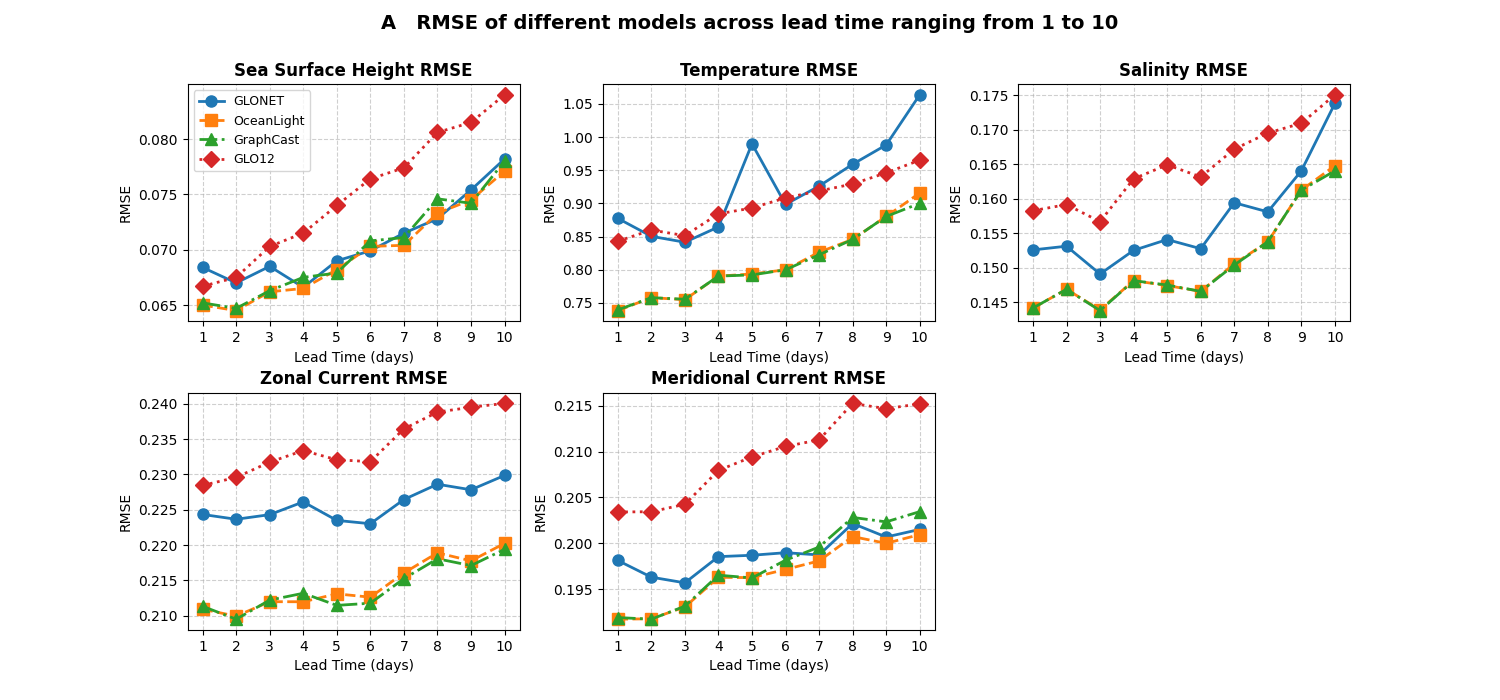}
    \end{subfigure}
    \begin{subfigure}{\textwidth}
        \centering
        \includegraphics[width=\textwidth]{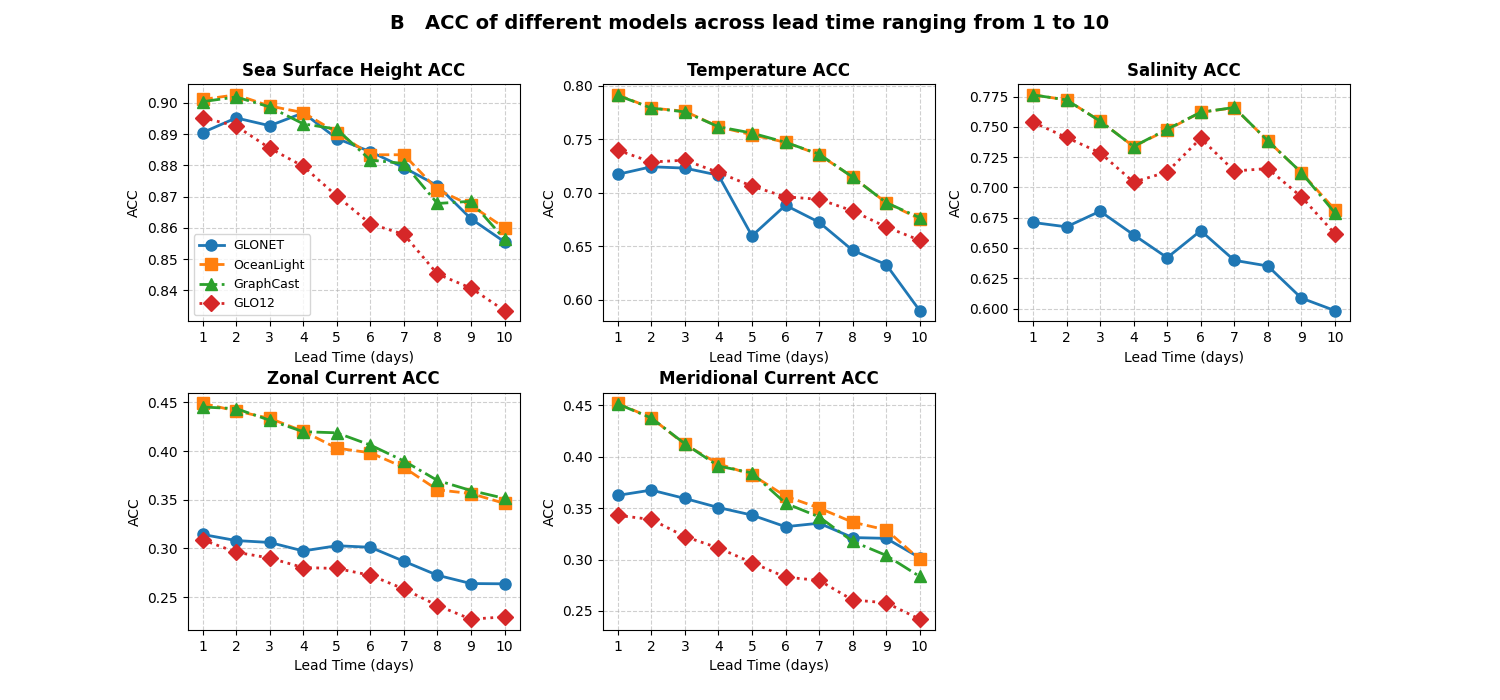}
    \end{subfigure}
    \caption{\textbf{Forecast accuracy of ocean state variables across GLONET, OceanLight, GraphCast, and GLO12.}
Root-mean-square error (RMSE) and anomaly correlation coefficient (ACC) are evaluated for five key ocean
variables as a function of forecast lead time (1--10 days).
\textbf{(A)} RMSE as a function of lead time for sea surface height (top left), temperature (top middle),
salinity (top right), zonal current (bottom left), and meridional current (bottom right). Lower RMSE indicates higher forecast accuracy.
\textbf{(B)} ACC as a function of lead time for the same five variables, arranged in the same layout as in (a). Higher ACC indicates stronger correspondence between forecast and observed anomalies. Data are shown as mean values for each model across the forecast period of 2024.}
    \label{fig:rmseacc}
\end{figure}

Overall, our model achieves consistently lower RMSE and higher ACC than GLONET across all variables and lead times, while showing performance comparable to or slightly better than GraphCast in most cases. For sea surface height (SSH), our model obtains the lowest RMSE at nearly all forecast horizons and maintains competitive ACC values, especially at short- and medium-range lead times. In temperature prediction, our method significantly reduces RMSE compared with GLONET and GLO12 and achieves similar ACC performance to GraphCast, indicating better stability during long-term forecasting. For salinity, both RMSE and ACC results demonstrate that our model substantially outperforms GLONET, GLO12 and closely matches GraphCast, particularly at 1–7 day lead times. In the prediction of zonal and meridional currents, our model consistently achieves lower RMSE and higher ACC than GLONET and GLO12, while remaining competitive with GraphCast. Although the forecasting accuracy of all models gradually decreases as the lead time increases, our method shows a slower degradation trend, especially for dynamic variables such as ocean currents. These results demonstrate that the proposed model has strong forecasting capability and generalization performance on multiple ocean variables under the IV-TT evaluation framework.

We also conducted a depth-stratified evaluation of temperature and salinity using the IVTT metric and plotted their vertical profiles. The vertical profiles of temperature and salinity are shown in supplementary Figure S1. We have also implemented the IVTT evaluation of ocean temperature, salinity, current velocity, and sea surface height, broken down by season and region. The experimental results are presented in the figures in supplementary Figure S2-S3.

\subsection{Increased dynamical consistency between geostrophic and surface flows
}
The geostrophic surface currents were diagnosed from the model-predicted sea surface height above geoid (SSH) via the geostrophic balance equations and compared with the model-predicted surface currents. Dynamical consistency between the two was quantified using the RMSE and correlation coefficient; lower RMSE and higher correlation coefficients indicate better geostrophic consistency.
The geostrophic surface currents (\(u_g\) and \(v_g\)) were derived from the model-predicted sea surface height above geoid; the calculation details are given in the supplementary note 1.

The comparison results are presented in figure 3.
\begin{figure}[H]
    \centering
    \includegraphics[width=\textwidth]{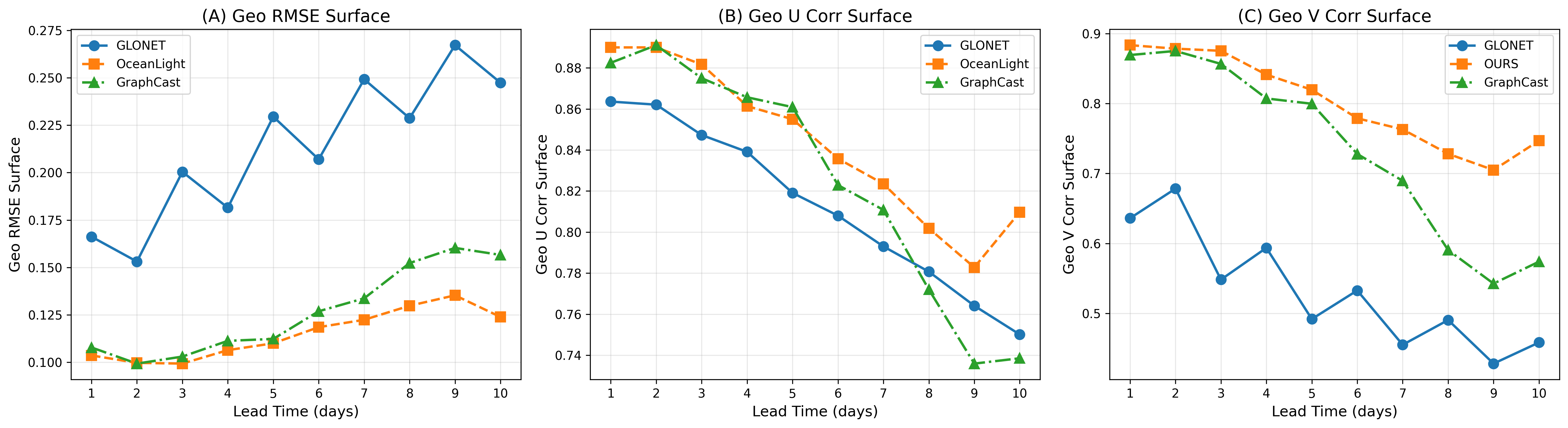}
    \caption{Internal geostrophic consistency of GLONET, OceanLight, and GraphCast surface current forecasts.
For each model, surface geostrophic currents are derived from the forecasted sea surface height(SSH) field and compared against the model's directly forecasted surface zonal($u$) and meridional ($v$) velocity components, as a function of forecast lead time (1-10 days).
\textbf{(A)} Root-mean-square error (RMSE) between SSH-derived geostrophic surface currents and forecasted surface currents (Geo RMSE Surface) as a function of lead time.
\textbf{(B)} Correlation coefficient between SSH-derived geostrophic zonal current and forecasted zonal current (Geo U Corr Surface) as a function of lead time.
\textbf{(C)} Correlation coefficient between SSH-derived geostrophic meridional current and forecasted meridional current (Geo V Corr Surface) as a function of lead time.
These metrics quantify the degree to which each model's forecasted surface velocity fields are dynamically consistent with its own forecasted SSH field under geostrophic balance.}
    \label{fig:internal}
\end{figure}

The experimental results show that the our model achieves the lowest RMSE between the geostrophic currents and the predicted surface currents, indicating that the predicted dynamical height field is more physically reasonable. Since geostrophic currents are directly determined by the spatial gradients of sea surface height, the lower RMSE suggests that the proposed method produces more accurate SSH gradient structures and reconstructs more realistic ocean dynamical patterns, rather than merely reducing point-wise prediction errors.

In addition, our model achieves the highest correlation coefficients for both the zonal and meridional velocity components, demonstrating stronger consistency in the flow direction between the derived geostrophic currents and the predicted surface currents. In particular, the improvement in the meridional component (\(v\)) is especially significant, indicating that the model more effectively captures the two-dimensional spatial structure of SSH gradients and mesoscale ocean dynamical features.

These results demonstrate that the proposed model not only fits sea surface height fields more accurately, but also better preserves the underlying ocean dynamical consistency and geostrophic balance relationships.

\subsection{Improved energy consistency and realistic kinetic energy distribution}
To further evaluate the dynamical characteristics of the predicted ocean surface circulation, the total kinetic energy (KE), mean kinetic energy (MKE), and eddy kinetic energy (EKE) were calculated based on the predicted velocity components (\(u,v\)) of 0.49m, and the calculation formulas are provided in the supplementary note 5.

We use the AVISO dataset as the observational reference to calculate KE, MKE and EKE. The same quantities are also computed from the model-predicted surface velocity fields. To quantitatively evaluate the consistency between the model-derived and AVISO-derived energy fields, several statistical metrics are calculated, including the RMSE, variance ratio, amplitude ratio, and area integral ratio, with the computational formulas presented in the supplementary note 3. We demonstrate the KE results and its visualization in figure 4. The results of EKE, MKE are demonstrated in supplementary S14-S15.
\begin{figure}[H]
    \centering

    \begin{subfigure}{\textwidth}
        \centering
        \includegraphics[width=\textwidth]{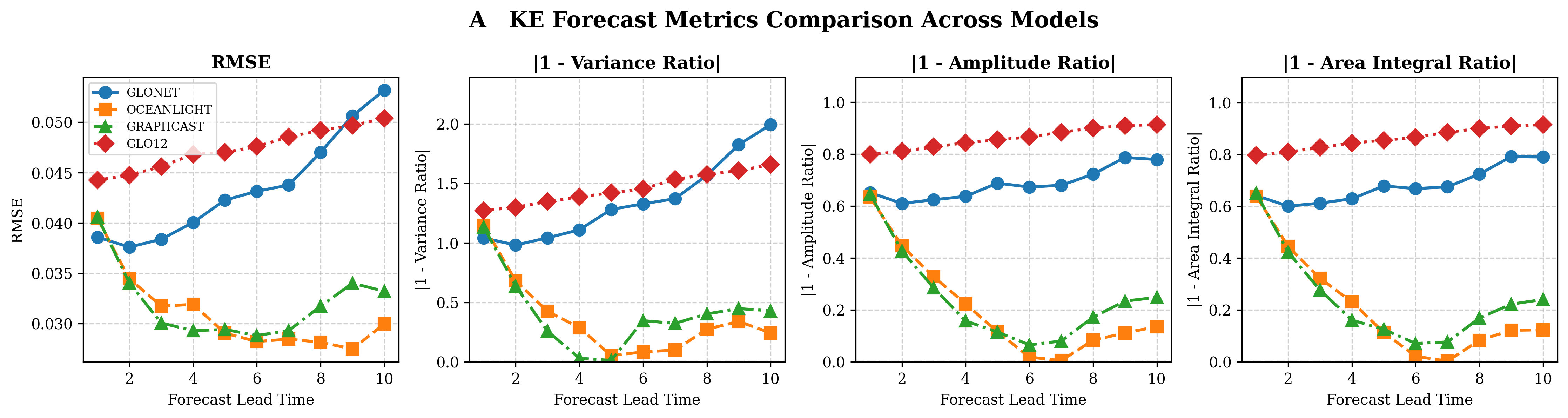}
    \end{subfigure}

    \vspace{0.3cm}

    \begin{subfigure}{\textwidth}
        \centering
        \caption*{\textbf{B\quad Comparison of 10-day forecast sea surface kinetic energy}}
        \includegraphics[width=\textwidth]{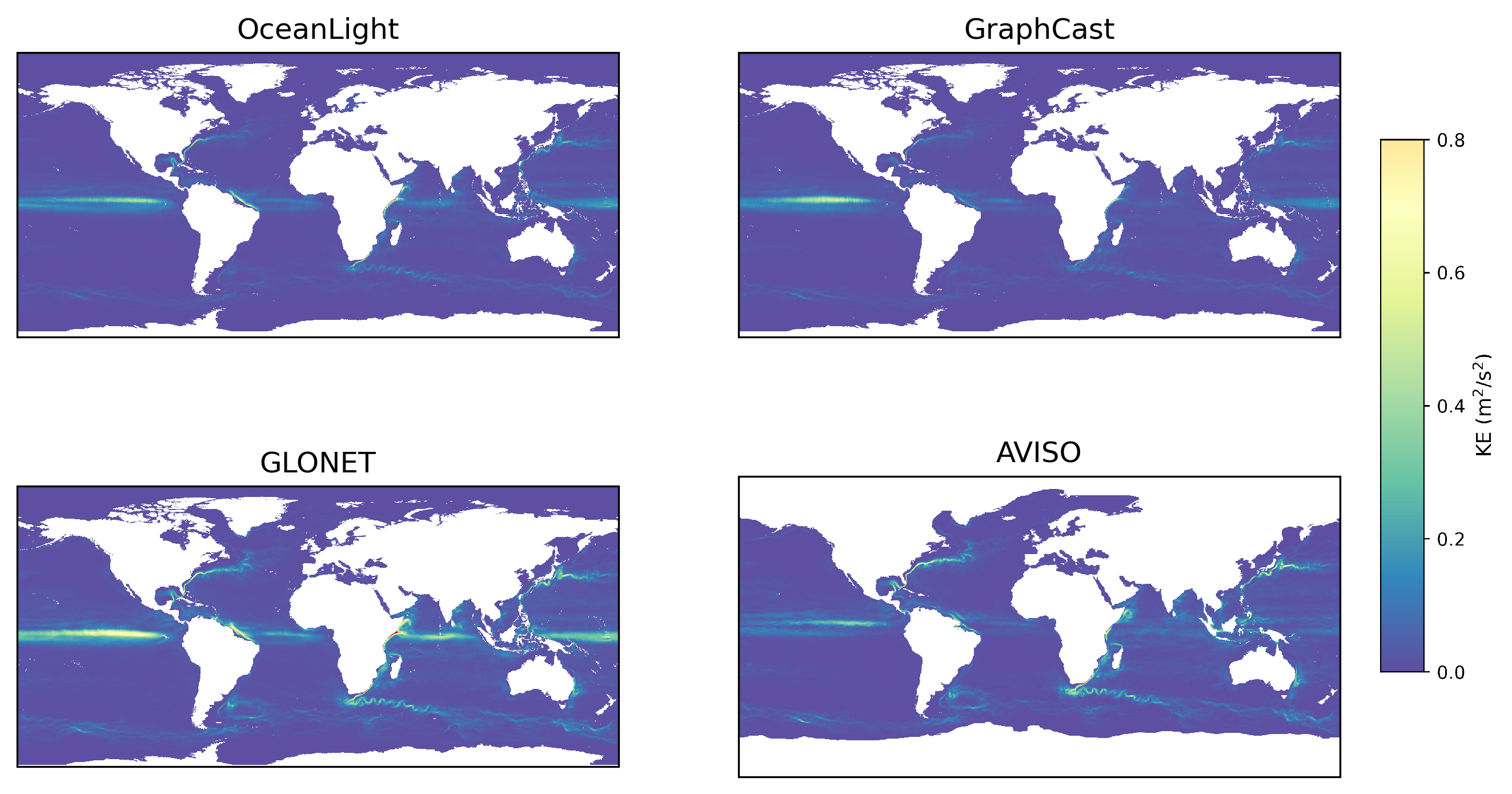}
    \end{subfigure}

    \caption{Comparison of model-derived and AVISO-derived surface kinetic energy fields across forecast lead times. For each model, kinetic energy (KE) is computed from the forecasted surface velocity fields (KE $= \frac{1}{2}(u^2+v^2)$, in m$^2$/s$^2$) and compared against the corresponding energy field derived from AVISO altimetry. \textbf{(A)} KE forecast skill as a function of forecast lead time, evaluated using four metrics: root-mean-square error (RMSE, leftmost panel) and three ratio-based metrics (right three panels), namely $|1-\text{Variance Ratio}|$, $|1-\text{Amplitude Ratio}|$, and $|1-\text{Area Integral Ratio}|$, expressed as absolute deviations from unity such that lower values indicate closer agreement with the AVISO-derived reference field. Four models are compared: GLONET, OceanLight, GraphCast, and GLO12. \textbf{(B)} Spatial maps of 10-day-averaged sea surface kinetic energy from three models—OceanLight, GraphCast, and GLONET—initialized using the OceanBench input dataset. The displayed fields represent the mean of each model’s 10-day forecasts, and are compared against AVISO satellite altimetry-derived observations.}
    \label{fig:energycombined}
\end{figure}

For RMSE, our model achieves the lowest errors across KE, MKE, and EKE, demonstrating substantially improved prediction accuracy compared with both GLONET and GraphCast. For the ratio-based metrics, including variance ratio, amplitude ratio, and area integral ratio, values closer to 1 indicate better agreement with the reference data. Variance ratio represents the magnitude of spatial variation in an energy field, a value smaller than 1 indicates overly smooth predictions, while a value larger than 1 suggests exaggerated spatial fluctuations. The amplitude ratio reflects whether the average energies of the two fields are matched, and the area integral ratio indicates whether the total energies of the two fields are consistent. The results show that our model achieves most of the ratio-based metrics closer to 1, indicating better agreement with the reference fields. In contrast, GLONET exhibits substantial overestimation of energy intensity and spatial variability, as all ratio-based metrics are significantly larger than 1. This suggests that GLONET tends to excessively amplify the geostrophic energy fields and produces overly exaggerated spatial fluctuations.

\subsection{Evaluation of Mesoscale Eddy Detection}
To evaluate the mesoscale eddy forecasting performance, the predicted Sea Surface Height above Geoid (SSH) was first converted into Sea Level Anomaly (SLA) by subtracting the Mean Dynamic Topography (MDT). Mesoscale eddies were then identified from the resulting SLA fields using the Py-Eddy-Tracker (PET) algorithm. As the reference, mesoscale eddies were extracted from the AVISO satellite-observed SLA using the same PET algorithm, ensuring a consistent eddy identification procedure for both forecasts and observations. For eddy-based verification, a forecasted eddy was matched to an observed eddy when either of the following criteria was satisfied: the core position of the forecasted eddy fell within the effective closed contour of the observed eddy, or the distance between the forecasted eddy core and the observed eddy core was less than or equal to the effective radius of the observed eddy. Based on the matched forecasted and observed eddies, four commonly used dichotomous verification metrics, namely Probability of Detection (POD), False Alarm Ratio (FAR), Bias, and Critical Success Index (CSI)~\cite{Metrics}, were calculated to comprehensively evaluate the detection capability, reliability, systematic prediction tendency, and overall forecasting skill of our model and the baseline methods. The calculation formulas are provided in supplementary note 4. Since the AVISO altimetry product used as the observational reference for eddy detection has a spatial resolution of 0.25°, the mesoscale eddy detection comparison is restricted to baselines at matching resolution, namely GLONET and GraphCast. GLO12 is excluded from this comparison due to its higher resolution of 1/12°, as GLO12 resolves finer-scale eddies that fall below what AVISO can detect, making direct comparison against the 0.25° reference inherently unfair to its true eddy-resolving skill.The results are shown in figure 5.

Western boundary current regions are among the most dynamically active areas of the global ocean. The Kuroshio Extension, Gulf Stream, and Brazil Current are characterized by strong currents, sharp sea surface height gradients, intense baroclinic instability, and vigorous air–sea interactions. These processes favor the frequent generation, propagation, and deformation of mesoscale eddies, making such regions rich in eddy activity but also particularly challenging for ocean forecasting. Therefore, in addition to global quantitative evaluation, we further conduct regional visualization and IVTT-based assessment~\cite{IVTT} over these three representative western boundary current regions to examine whether the model can accurately capture mesoscale eddy structures and SLA evolution in highly energetic oceanic regimes.

Thus, mesoscale eddy visualizations were conducted in three major western boundary current regions, including the Kuroshio and its extension region, the Gulf Stream, and the Brazil Current. The initial condition was taken from the OceanBench dataset on 2 January 2024, and a 10-day lead-time forecast was performed. The AVISO observational data on 12 January 2024 were used as the reference state. Mesoscale eddies were identified from both the forecasted and observed SLA fields using the PET algorithm~\cite{pet}, and the corresponding detection results are shown in supplementary Figure S4-S6.
\begin{figure}[H]
    \centering   \includegraphics[width=\textwidth]{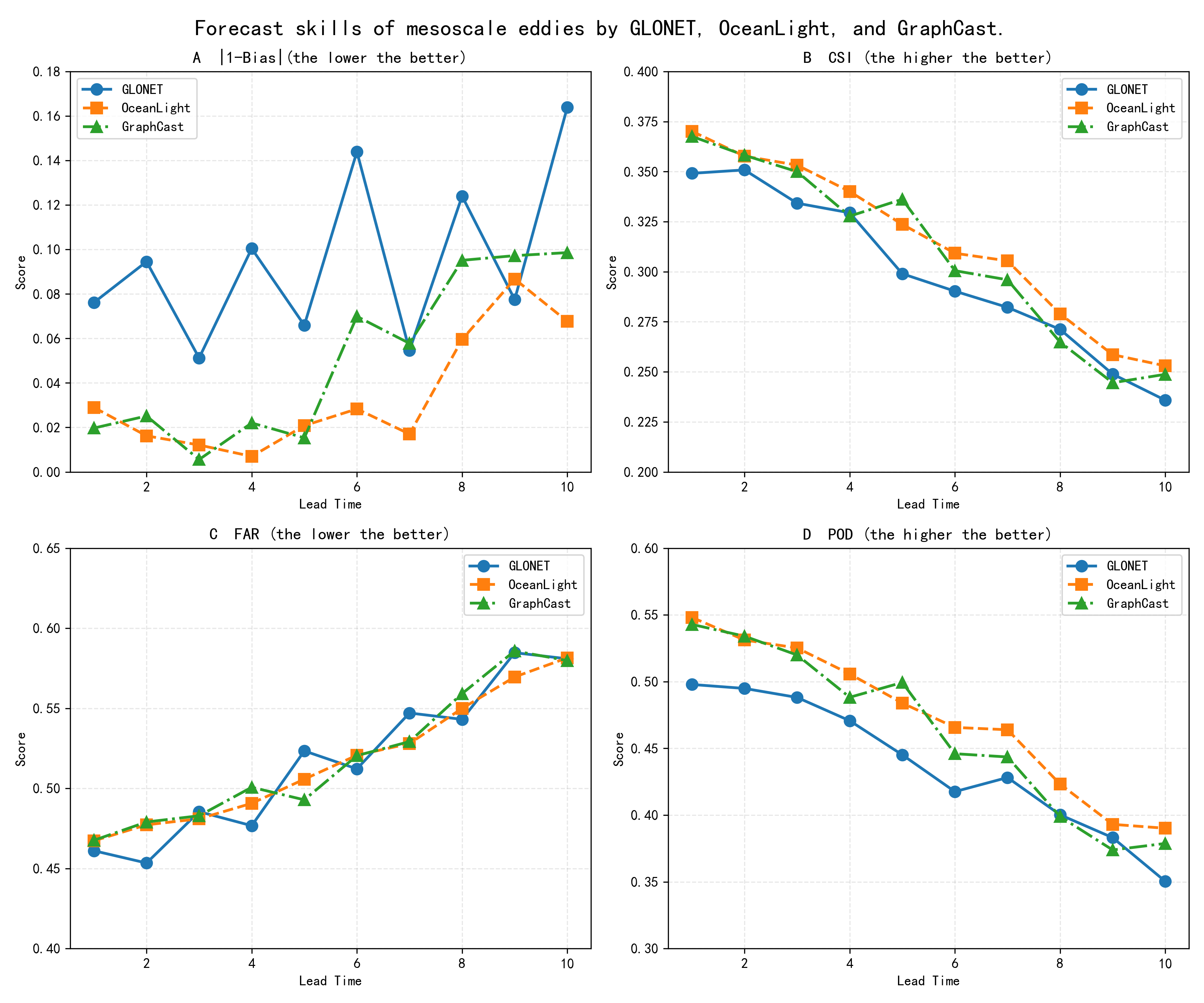}
    \label{fig:eddy_cal}
    \caption{Verification of forecasted mesoscale eddies against AVISO altimetry across forecast lead times. For each lead time, forecasted sea surface height (SSH) fields from three forecast systems were converted to sea level anomaly (SLA) by subtracting the mean dynamic topography (MDT), and mesoscale eddies were then identified from the resulting SLA fields using the py-eddy-tracker (PET) algorithm. The same detection procedure was applied to satellite-observed AVISO SLA fields, which serve as the reference for eddy identification. Forecast-detected eddies at each lead time were matched against the AVISO reference eddies, and four
    categorical verification metrics were computed as a function of lead time (1-10) for GLONET (blue circles), our model (OceanLight; orange squares) and GraphCast (green triangles). \textbf{(A)} Absolute deviation of the eddy-count bias ratio from unity, $|1-\mathrm{Bias}|$, where Bias is the ratio of the number of forecast-detected eddies to the number of AVISO-detected eddies; values near zero indicate agreement in eddy abundance between forecast and observations. \textbf{(B)} Critical success index (CSI), which jointly penalizes missed and falsely detected eddies; higher values indicate better overall detection skill. \textbf{(C)} False alarm ratio (FAR), the fraction of forecast-detected eddies with no matching observed eddy; lower values are better. \textbf{(D)} Probability of detection (POD), the fraction of AVISO-observed eddies correctly recovered by the forecast; higher values are better.}
\end{figure}

\subsection{Enabling efficient and scalable ocean forecasting with reduced computational demands}
To ensure a fair comparison with GraphCast, our unstructured mesh-based graph neural network model adopts identical hyperparameter settings, i.e., an embedding size of 512 and blocks number of 16.

\begin{table}[htbp]
\centering
\caption{Comparison of Nodes and Edges Between OceanLight and GraphCast}
\label{tab:mesh_comparison}
\begin{tabular}{lcccc}
\toprule
Model & Mesh nodes & Mesh edges & grid$\rightarrow$mesh edges & mesh$\rightarrow$grid edges \\
\midrule
Graphcast  & 40,962 & 2,941,920 & 1,539,243 & 2,941,920 \\
OceanLight       & 1212 & 4120  & 681,067   & 681,067   \\
\bottomrule
\end{tabular}
\end{table}

As shown in table 1, compared with GraphCast, our model reduces the numbers of mesh nodes and mesh edges by approximately 97.0\% and 99.9\%, respectively, and decreases the grid→mesh edges and mesh→grid edges by approximately 55.7\% and 76.8\%, respectively. 
The significant reduction in the number of edges and nodes greatly decreases both the FLOPs and the memory usage during model operation, as shown in table 2. Specifically, FLOPs are reduced by 73.0\%, while Max Memory Reserved and Max Memory Allocated are reduced by 62.4\% and 62.3\%, respectively. The decrease in FLOPs also greatly reduces the time cost of model training, and the time decomposition of model training is illustrated in the figure 6.

\begin{table}[htbp]
\centering
\caption{Comparison of FLOPs and max memory during training Between OceanLight and GraphCast}
\label{tab:mesh_comparison}
\begin{tabular}{lcccc}
\toprule
Model & Graphcast & OceanLight \\
\midrule
FLOPs  & 27673.207G & 7481.553G  \\
Max Memory Reserved & 77.70GB & 29.19 GB  \\
Max Memory Allocated & 69.09GB & 26.04GB  \\
\bottomrule
\end{tabular}
\end{table}

Compared to GraphCast, our model reduces the training time of the encoder, processor, decoder, and backpropagation by 40.3\%, 92.2\%, 61.4\%, and 71.2\%, respectively.
\begin{figure}[H]
    \centering  
    \includegraphics[width=\linewidth]{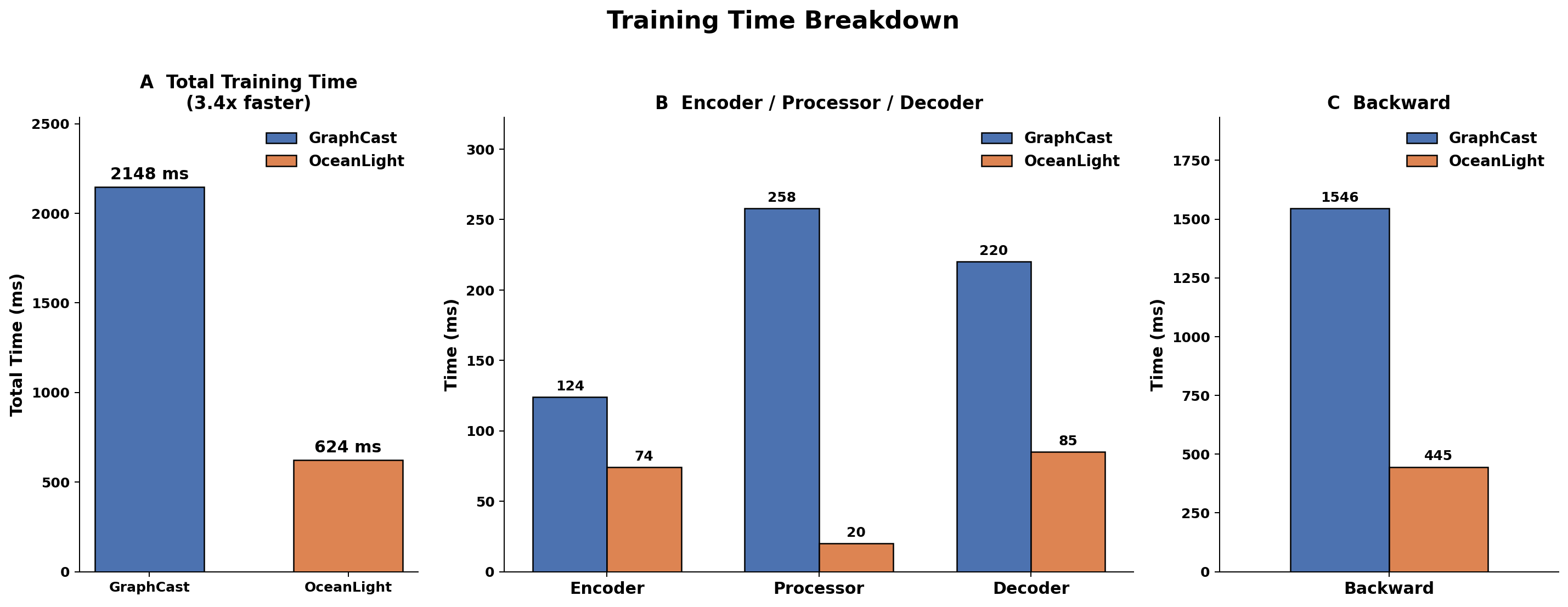}
    \caption{Training time comparison between GraphCast and OceanLight. \textbf{(A)} Total training time per iteration, showing an overall 3.4× speedup for OceanLight compared to GraphCast. (B) Per-component breakdown of the forward pass (Encoder, Processor, Decoder): OceanLight reduces Encoder time by 40.3\%, Processor time by 92.2\%, and Decoder time by 61.4\%, with the Processor showing the largest relative reduction. \textbf{(C)} Backward pass time, which dominates the total training cost for both models; OceanLight reduces backward time by 71.2\% compared to Graphcast.}
\end{figure}

\section{Methods}\label{sec3}
This section presents the methodology for constructing an unstructured mesh representation for global ocean forecasting at 0.25° resolution using Graph Neural Networks (GNNs). Figure 1 illustrates the overall methodological framework.

\subsection{Overall Forecasting Framework}
The core architecture follows the established encode-process-decode paradigm used in GraphCast. Given an input ocean state at time $t$, the model predicts the ocean state at time $t+\Delta t$. The key distinction lies in the graph topology: instead of uniform refinement to an icosahedral mesh, our model operates on unstructured mesh comprising 1,212 regions based on ocean climatology data. This results in a substantially sparser graph structure, with only 4,120 edges connecting mesh nodes, and 681,067 bidirectional edges linking grid and mesh nodes. Compared to GraphCast’s dense mesh, our unstructured representation reduces the number of mesh nodes and edges by over an order of magnitude, cutting GPU memory usage by over 50\% while preserving—and even slightly improving—the model’s capacity to resolve regionally varying ocean dynamics. Based on the graph structure constructed from the above process, we use 
$\mathbf{v}_{\scriptscriptstyle G}$, $\mathbf{v}_{\scriptscriptstyle M}$, $\mathbf{e}_{\scriptscriptstyle G2M}$, $\mathbf{e}_{\scriptscriptstyle M2G}$, and $\mathbf{e}_{\scriptscriptstyle M2M}$ 
to denote the features of grid nodes, mesh nodes, edges from grid to mesh, 
edges from mesh to grid, and edges from mesh to mesh, respectively. 

The encoder embeds grid nodes, mesh nodes, and grid-to-mesh edge features into a latent space using separate MLPs. Then it performs a gnn-based message-passing step from grid nodes to mesh nodes. In this step, the 0.25° ocean data is encoded into the unstructured mesh representation:
\[
\tilde{\mathbf{v}}_G = \mathrm{MLP}(\mathbf{v}_G), \quad
\tilde{\mathbf{v}}_M = \mathrm{MLP}(\mathbf{v}_M), \quad
\tilde{\mathbf{e}}_{G2M} = \mathrm{MLP}(\mathbf{e}_{G2M})
\]
\[
\mathbf{v}_M' = \tilde{\mathbf{v}}_M + \mathrm{InteractionNetwork}\left(\tilde{\mathbf{v}}_G,\ \tilde{\mathbf{v}}_M,\ \tilde{\mathbf{e}}_{G2M}\right)
\]
\[
\mathbf{v}_G' = \tilde{\mathbf{v}}_G + \mathrm{MLP}(\tilde{\mathbf{v}}_G)
\]

The processor consists of 16 unshared message-passing layers that operate on the mesh representation. Each layer updates edges and nodes sequentially, followed by residual connections for both nodes and edges. For layer $l = 1, \dots, L$ (unshared parameters), where $l$ denotes the index of the current processor layer among the $L$ stacked message-passing layers:
\[
\mathbf{e}_{M2M}^{(l)},\ \mathbf{v}_M^{(l)} = \mathrm{EdgeUpdate\&NodeUpdate}\left(\mathbf{v}_M^{(l-1)},\ \mathbf{e}_{M2M}^{(l-1)}\right)
\]

\[
\mathbf{v}_M^{(l)} \leftarrow \left(\mathbf{v}_M^{(l-1)} + \mathbf{v}_M^{(l)}\right), \quad
\mathbf{e}_{M2M}^{(l)} \leftarrow \mathbf{e}_{M2M}^{(l-1)} + \mathbf{e}_{M2M}^{(l)}
\]

\[
\mathbf{v}_M^{\mathrm{out}} = \mathbf{v}_M^{(L)}
\]

The decoder uses MLPs to embed mesh-to-grid edge features and performs gnn-based message-passing step from mesh nodes back to grid nodes. MLPs are used for the message passing flow, converting the information contained in the meshes to grids and output the prediction:
\[
\tilde{\mathbf{e}}_{M2G} = \mathrm{MLP}(\mathbf{e}_{M2G})
\]
\[
\mathbf{v}_G'' = \mathbf{v}_G' + \mathrm{InteractionNetwork}\left(\mathbf{v}_M^{\mathrm{out}},\ \mathbf{v}_G',\ \tilde{\mathbf{e}}_{M2G}\right)
\]
\[
\hat{\mathbf{v}}_G^{\mathrm{out}} = \mathrm{MLP}(\mathbf{v}_G'')
\]
\subsection{Data Preprocessing}
We obtained monthly ocean climatological data from the Copernicus Marine Environment Monitoring Service (CMEMS).
For each grid point, we extract a multi-dimensional feature vector comprising various oceanographic variables, including temperature, salinity, zonal and meridional ocean current velocity components at 23 discrete depth levels (i.e., 0.49m, 2.65m, 5.08m, 7.93m, 11.41m, 15.81m, 21.60m, 29.44m, 40.34m, 55.76m, 77.85m, 92.32m, 109.73m, 130.67m, 155.85m, 186.13m, 222.48m, 266.04m, 318.13m, 380.21m, 453.94m, 541.09m and 643.57m) and sea surface height. By averaging the climatological fields of all twelve months, we construct the annual mean ocean climatology. This dataset represents the long-term mean state of the ocean and reflects its stable background characteristics. The annual mean climatological dataset provides a robust representation of the ocean’s integrated mean state on long time scales, serving as the foundational dataset for subsequent calculation and ocean graph construction.

Global ocean data at high resolution contains fine-scale details that can introduce challenges for constructing a tractable and representative adaptive mesh. The high-resolution data may contain noise, and in regions such as land-sea boundaries or strong oceanic fronts, fine-scale spatial gradients can be excessively sharp, potentially leading to an overly fragmented adaptive mesh with an impractically large number of regions. To address these issues, we employ a hierarchical multi-resolution strategy that balances computational efficiency, noise reduction, and the preservation of meaningful spatial structure for mesh construction. We regridded the data from its original resolution to a 1.40625° × 1.40625° latitude-longitude grid using the Climate Data Operators (CDO)~\cite{cdo} with bilinear interpolation (remapbil operator).

Denote the horizontal grids of 1.40625° ocean climatology as
\[
  \Omega_c = \{(i,j) \mid i = 1, \ldots, N_{lat}^c,\ j = 1, \ldots, N_{lon}^c\}  
\]
At each grid point $(i,j)\in\Omega_c$, the multivariate climatological state is represented by a feature vector $\mathbf{x}_{i,j} \in \mathbb{R}^d$
which includes temperature, salinity, zonal and meridional velocities across multiple depth levels, as well as sea surface height. To extract dominant mode of variability, principal component analysis (PCA)~\cite{pca1, pca2} is applied to the multivariate feature vectors, reducing it to a one-dimensional principal component. Formally, the reduced scalar value at each grid point is given by
\[
z_{i,j}=\mathcal{P}(\mathbf{x}_{i,j}),\;z_{i,j}\in \mathbb{R}
\]
where $\mathcal{P}(\cdot)$ denotes projection onto the first principal component via PCA. The resulting scalar field is
\[
Z = \{z_{i,j}\}\in\mathbb{R}^{N_{lat}^c\times N_{lon}^c}
\]
Then the spatial gradient of $Z$ is computed using the sobel operator~\cite{sobel}. Let $\mathbf{K}_x$ and $\mathbf{K}_y$ denote the standard Sobel kernels in the zonal and meridional directions respectively, which are defined as:
\[
\mathbf{K}_x = \begin{bmatrix}
-1 & 0 & 1 \\
-2 & 0 & 2 \\
-1 & 0 & 1
\end{bmatrix}\qquad
\mathbf{K}_y=\begin{bmatrix}
-1 & -2 & -1 \\
0 & 0 & 0 \\
1 & 2 & 1
\end{bmatrix}
\]
The discrete gradients at each grid point $(i,j)$ are calculated as:
\[
G_x(i, j) = (\mathbf{K}_x * Z)(i, j) = \sum_{p=-1}^{1} \sum_{q=-1}^{1} K_x(p, q) Z(i + p, j + q)
\]
\[
G_y(i, j) = (\mathbf{K}_y * Z)(i, j) = \sum_{p=-1}^{1} \sum_{q=-1}^{1} K_y(p, q) Z(i + p, j + q)
\]
The gradient magnitude is then given by
\[
G(i,j)=\sqrt{G_x(i,j)^2+G_y(i,j)^2},
\]
This resulting in the gradient map $G\in\mathbb{R}^{N_{lat}^c\times N_{lon}^c}$, 
and the gradient field is used exclusively to guide the selection of seed points in the subsequent region-growing procedure.
\subsection{Region Growing Partition} 
Inspired by work of adaptive multi-granularity graph
representation~\cite{adaptive}, we first adopt the region growing method for initial region delineation. For any pair of adjacent grid points $(i,j)$ and $(m,n)$, the feature-space difference is defined as
\[
D((i,j),(m,n)) = \|\mathbf{x}_{i,j} - \mathbf{x}_{m,n}\|_2
\]
Let $\mathcal{N}(i,j)$ denote the 8-connected neighborhood of $(i,j)$, $\tau$ denote the threshold as the domain is partitioned into spatially connected regions using the threshold-based region-growing algorithm. Let $\Omega \subseteq \Omega_u$ denote the set of unlabeled grid points. At each iteration, a new region is initialized from the point with the minimum gradient magnitude:
\[
(i_k, j_k) = \arg \min_{(i,j) \in \Omega_u} G(i,j).
\]
Starting from this seed, the region $R_k$ is expanded by recursively including neighboring points that satisfy the similarity criterion:
\[
D((i,j),(m,n)) \leq \tau, \quad (m,n) \in \mathcal{N}(i_k,j_k).
\]
The expansion is implemented using depth-first traversal until no further points can be added. This process is repeated until all grid points are assigned, yielding a partition
\[
\Omega_c = \bigcup_{k=1}^{K} R_k, \qquad R_k \cap R_\ell = \varnothing \ (k \neq \ell).
\]
While the region growing algorithm preliminary partitioning method effectively handles regions such as land-sea boundaries and areas with steep gradients, it still generates large mesh regions that exceed over 20\% of the total ocean grid points. Such large regions are problematic because they aggregate diverse oceanic features into a single mesh node, reducing the model's ability to resolve spatial heterogeneity and potentially degrading forecast skill.

\subsection{HDBSCAN Clustering} 
To further resolve large-scale structures, regions occupying more than 20\% of the domain are refined using HDBSCAN clustering. For each region $R_k$, define its area fraction:
\[
\alpha_k = \frac{|R_k|}{|\Omega_c|}.
\]
Regions satisfying $\alpha_k>0.20$ are selected for refinement. For each such region, feature vectors $\mathbf{x}_{i,j}$ in the region $R_k$ forms a sample set\[
X_k=\{\mathbf{x}_{i,j}\mid (i,j)\in R_k\}
\]
HDBSCAN partitions \( X_k \) into clusters, inducing a decomposition
$R_k = \bigcup_{l=1}^{L} R_{k,l}$,
where each \( R_{k,l} \) denotes a refined subregion.
Since HDBSCAN operates in feature space, a given cluster $R_{k,l}$ may consist of multiple geographically disconnected components. To ensure spatial consistency, each cluster is further decomposed into spatially connected subregions based on grid adjacency. Specifically, for each $R_{k,l}$, we define its connected components as
\[
R_{k,l} \;\longrightarrow\; \{R_{k,l}^{(1)}, R_{k,l}^{(2)}, \dots\},
\]
where each $R_{k,l}^{(\ell)}$ is a maximal connected subset under the neighborhood relation $\mathcal{N}(i,j)$.
These connected components are treated as independent regions in the final segmentation, and subclusters replace the original region, yielding a refined segmentation of the global ocean.

\subsection{Mesh Graph Construction} Following the segmentation, each region $R_k$ is treated as a mesh node. The set of mesh nodes is defined as
\[
\mathcal{V}_M=\{R_1, R_2,\dots,R_k\}
\]
Each mesh node $R_k$ is associated with a representative location in the spatial domain, defined as the centroid of the region:
\[
(\bar{i}_k, \bar{j}_k) = \frac{1}{|R_k|} \sum_{(i,j) \in R_k} (i, j).
\]
Two meshes $R_k$ and $R_\ell$ are connected if they share at least one pair of neighboring grid points:
\[
(R_k, R_\ell) \in \mathcal{E}_1 \iff \exists(i, j) \in R_k, (m, n) \in R_\ell \text{ s.t. } (m, n) \in \mathcal{N}(i, j).
\]
Here, $\mathcal{E}_1$ denotes the set of first-order geographical adjacency edges. To capture higher-order spatial relationships, edges are further extended based on multi-step connectivity. Let\[
\mathcal{G}_1=(\mathcal{V}_M, \mathcal{E}_1)
\]
denote the first-order adjacency graph.
For a given integer $k\geq1$, define\[
\mathcal{E}_M = \{(R_i, R_j) \mid \text{dist}_{\mathcal{G}_1}(R_i, R_j) \leq k,\; R_i \neq R_j \},
\]
where $\text{dist}_{\mathcal{G}_1}$ is the shortest-path distance. In this study, we set $k=2$.
The final mesh graph is defined as\[
\mathcal{G}_M=(\mathcal{V}_M, \mathcal{E}_M)
\]
where $\mathcal{E}_M$ includes both first-order and higher-order (k-hop) spatial connectivity.

\subsection{Interaction Network}
Let
\[
\Omega_f=\{(u,v)\mid u=1,\dots,N_{lat}^f,v=1,\dots,N_{lon}^f\}
\]
denote the fine-resolution spatial grid at 
0.25°, on which the subsequent gnn forecasting model is applied. The set of fine-grid nodes is defined as
\[
\mathcal{V}_G=\Omega_f
\]
Since the mesh partition is obtained on the coarse grid, each 0.25\textdegree grid point is assigned to a mesh region based on its spatial location. Specifically, a 0.25\textdegree grid node $(u,v)$ is associated with region $R_k$ if its geographical coordinates fall within the spatial extent of $R_k$.
Let
\[
\phi : \Omega_f \rightarrow \mathcal{V}_M
\]
denote this assignment, where $\phi(u,v)=R_k$ if the grid point $(u,v)$ lies within the spatial domain covered by region $R_k$. The grid–mesh edge set is then defined as
\[
\mathcal{E}_{GM} = \{((u, v), R_k), (R_k, (u, v)) \mid \phi(u, v) = R_k\}.
\]
Thus, for each 0.25\textdegree grid node $(u,v)$, two directed edges are constructed:
\[
(u,v)\rightarrow R_k,\qquad R_k \rightarrow (u,v),\qquad \text{if } \phi(u,v)=R_k
\]
thereby enabling bidirectional information exchange between grid nodes and mesh nodes.

The complete graph is defined as
\[
\mathcal{G}=(\mathcal{V}, \mathcal{E}),
\]
where\[
\mathcal{V} = \mathcal{V}_G \cup \mathcal{V}_M, \quad \mathcal{E} = \mathcal{E}_M \cup \mathcal{E}_{GM}
\]
Given the constructed edge set $\mathcal{E}_{GM}$, information is then propagated between nodes via an Interaction Network. For an arbitrary directed edge $(s,r) \in \mathcal{E}$, where $s \in \mathcal{V}$ denotes the sender node and $r \in \mathcal{V}$ the receiver node, let $x_s$ and $x_r$ denote the corresponding node features and $e_{sr}$ the edge feature. In the Interaction Network, the receiver node representation is updated through message passing, followed by aggregation and node update.

As for the message passing step, for each edge $(s,r) \in \mathcal{E}$, a message is computed by concatenating the sender feature, the receiver feature, and the edge feature, followed by a multi-layer perceptron:
\[
    e'_{sr} = \mathrm{MLP}\left( \left[ x_s \, \Vert \, x_r \, \Vert \, e_{sr} \right] \right),
\]
where $[\cdot \Vert \cdot \Vert \cdot]$ denotes feature concatenation.

As for the aggregation and node update step, for each receiver node $r$, the messages from all its incoming edges are aggregated by summation, and the receiver representation is updated via a second multi-layer perceptron:
\[
    x'_r = \mathrm{MLP}\left( \left[ x_r \, \Vert \sum_{s \, : \, (s,r) \in \mathcal{E}} e'_{sr} \right] \right).
\]
This topological design of grid-mesh edges reduces the number of edges while preserving the essential edges for physical information transmission. Thus, the enhancement in computational efficiency is achieved without compromising forecasting performance.

\section{Discussion}\label{sec4}

In this work, we present a lightweight ocean forecasting framework based on unstructured mesh representation, which is built according to ocean climatology. The construction of the unstructured mesh integrates the region growing method and the HDBSCAN clustering approach, enabling a relatively coarse mesh representation in stable and gentle open ocean areas, while employing a finer mesh in regions with drastic changes, such as land-sea boundaries. This mesh generation strategy effectively balances simulation accuracy and computational efficiency. Building on this efficient spatial representation, our framework achieves high forecasting accuracy and maintains high physical consistency with regard to geostrophic flow balance and kinetic energy fidelity, while significantly reducing memory consumption and time overhead compared to GraphCast and GLONET. The method also demonstrates the ability to identify and generate mesoscale eddies, indicating that the reduction in computational cost does not adversely affect its eddy-resolving capability. 

The reduction in GPU memory usage during training allows our model to be trained on a single 40GB GPU, in contrast to GLONET, which requires multiple 40GB GPUs and a model-parallel training strategy. In addition to the reduced memory footprint, our model's better time efficiency further contributes to a significant decrease in training cost.

Despite these advantages, OceanLight does have some limitations. The unstructured mesh used in this model is constructed based on climatological state, meaning that it primarily reflects long-term averaged oceanic conditions rather than capturing daily dynamical variability. Consequently, the mesh does not account for transient, short-term ocean processes that could be relevant for high-resolution forecasting. Future research will investigate methods for rapidly constructing unstructured meshes according to daily ocean conditions, thereby enabling more efficient and accurate ocean forecasting. Beyond the mesh construction strategy, another limitation lies in the forecasting paradigm itself: the current forecasting system is deterministic and does not provide probabilistic predictions. Leveraging the substantial reduction in computational cost afforded by the unstructured mesh representation, future work will further explore ensemble forecasting strategies, thereby enabling reliable uncertainty quantification for operational ocean prediction.

Looking beyond these immediate directions for improvement, we emphasize that the core contribution of this work extends beyond the specific GNN implementation evaluated here: the proposed unstructured mesh representation strategy constitutes an architecture-agnostic spatial representation for ocean state forecasting. Because the unstructured mesh is inherently sparse and irregular, each cell-level token interacts with only a geometrically local neighborhood, making the representation directly compatible with sparse transformer architectures~\cite{sparsetransformer, gencast} and other attention-based models that exploit irregular token connectivity. This decoupling of spatial discretization from the model backbone establishes a generalizable paradigm: future work may substitute the GNN with sparse self-attention mechanisms, neural operators, or hybrid physics–data-driven architectures~\cite{PINN, hybrid, Ocean-E2E} without modifying the underlying mesh representation. By framing the ocean state as an unstructured token graph rather than a structured image, this approach reconciles the computational efficiency demonstrated in our experiments with the physical fidelity—captured through eddy-resolving skill and dynamical consistency—required for scientific credibility, opening a new direction for scalable, physics-aware data-driven oceanography.

\section{Data availability}
We download reanalysis data from the official website of Copernicus Marine Service at \url{https://data.marine.copernicus.eu/product/GLOBAL_MULTIYEAR_PHY_001_030/download}, and the Aviso satellite altimetry observations at \url{https://data.marine.copernicus.eu/product/SEALEVEL_GLO_PHY_CLIMATE_L4_MY_008_057/download}. We download input datasets for OceanBench challenger evaluation following the document of Oceanbench at \url{https://oceanbench.readthedocs.io/en/latest/input-datasets-for-oceanbench-challenger-evaluation.html}.

\section{Code availability}
The mesh construction code is available at \url{https://github.com/Rowena-929/OceanLight}. The GNN-based forecasting model is implemented using an open-source PyTorch reimplementation of GraphCast, available at \url{https://github.com/openclimatefix/graph_weather}. Our complete codebase, which integrates the proposed unstructured mesh representation with the above GraphCast-style architecture, is hosted at \url{https://github.com/Rowena-929/OceanLight}.

\section*{Acknowledgements}
This research is partially supported by National Key R\&D Program of China (2024YFC3109200), National Natural Science Foundation of China (Grant No. 62372460), Hunan Provincial Natural Science Foundation of China (2024JJ4042), and the science and technology innovation Program of Hunan Province (2024RC3134).

\section*{Author Contributions Statement}
X.W and W.W designed the project. W.W performed model training. W.W and Q.Y.M performed model evaluation. W.W, J.X.Z, X.W., H.Z.L. and J.Q.S. wrote and revised the manuscript.

\section*{Competing interests}
The authors declare no competing interests.

\clearpage

\bibliography{sn-bibliography}
\clearpage

\renewcommand{\thefigure}{S\arabic{figure}}
\setcounter{figure}{0}
\section*{Supplementary}

\subsection*{Supplementary Note 1-Geostrophic current calculation}
\label{supp:sec:geostrophic}
The geostrophic surface currents ($u_g$ and $v_g$) were derived from the model-predicted sea surface height above geoid (SSH, denoted here as $\eta$) based on the geostrophic balance equations:

\[
u_g = -\frac{g}{f}\frac{\partial \eta}{\partial y}
\label{eq:supp:ug}
\]

\[
v_g = \frac{g}{f}\frac{\partial \eta}{\partial x}
\label{eq:supp:vg}
\]

where $g$ is the gravitational acceleration, $f$ is the Coriolis parameter, 
and $\partial \eta/\partial x$ and $\partial \eta/\partial y$ are the zonal 
and meridional gradients of SSH.

The geostrophic and model-predicted current speeds were calculated as

\[
S_g = \sqrt{u_g^2 + v_g^2}
\label{eq:supp:sg}
\]

\[
S_m = \sqrt{u^2 + v^2}
\label{eq:supp:sm}
\]
\;

\subsection*{Supplementary Note 2-Basic Evaluation Metrics}
\textbf{RMSE (Root Mean Square Error)} \newline
We use RMSE to measure pointwise prediction accuracy. It is formulated as the following:
\[
\mathrm{RMSE} = \sqrt{\frac{1}{N} \sum_{i=1}^{N} (\hat{y}_i - y_i)^2}
\]
where $N$ denotes the number of grids, $\hat{y}_i$, denotes the predicted value, and $y_i$ denotes the real value.

\textbf{ACC(Anomaly Correlation Coefficient)} \newline
We use ACC to assess the spatial pattern correlation with respect to climatological anomalies. It is formulated as the following:
\[
ACC = \frac{\sum_{i=1}^N (\hat{y}_i - c_i)(y_i - c_i)}{\sqrt{\sum_{i=1}^N (\hat{y}_i - c_i)^2} \sqrt{\sum_{i=1}^N (y_i - c_i)^2}}
\]
where $N$ denotes the number of grids, $\hat{y}_i$, denotes the predicted value, and $y_i$ denotes the real value, and $c_i$ denotes the corresponding climatology value.

\textbf{Corr(Correlation Coefficient)} \newline
We use the Pearson correlation coefficient (Corr) to evaluate the consistency between the geostrophic currents derived from sea surface height and the predicted surface currents. The correlation coefficients are calculated separately for the zonal and meridional velocity components. It is formulated as follows:

\[
\mathrm{Corr} =
\frac{
\sum_{i=1}^{N}
(\hat{u}_i - \overline{\hat{u}})
(u_i - \overline{u})
}{
\sqrt{
\sum_{i=1}^{N}
(\hat{u}_i - \overline{\hat{u}})^2
}
\sqrt{
\sum_{i=1}^{N}
(u_i - \overline{u})^2
}
}
\]

where \(N\) denotes the number of grids, \(\hat{u}_i\) denotes the geostrophic current component derived from sea surface height, \(u_i\) denotes the corresponding predicted surface current component, and \(\overline{\hat{u}}\) and \(\overline{u}\) denote their spatial mean values, respectively. The same formulation is applied to the meridional velocity component.\newline

\subsection*{Supplementary Note 3-Energy Evaluation Metrics}
To quantitatively evaluate the consistency between the model-predicted and observation-derived ocean surface kinetic energy fields, three spatial ratio metrics are applied to each energy variable — total kinetic energy~(KE), mean kinetic energy~(MKE), and eddy kinetic energy~(EKE). These metrics assess different aspects of energy agreement: the spatial variability amplitude, the mean energy intensity, and the total integrated energy magnitude, respectively.  For a
given energy field $X \in \{\mathrm{KE},\,\mathrm{MKE},\,\mathrm{EKE}\}$, the
three metrics are defined as follows.\newline
\textbf{Variance ratio} \newline
We use the variance ratio to evaluate whether the spatial variability amplitude of the model-predicted field is realistically reproduced. It is formulated as follows:
\[
\mathrm{Variance\ Ratio}(X)
    = \frac{\mathrm{Var}(X_{\mathrm{model}})}
                  {\mathrm{Var}(X_{\mathrm{obs}})}
           = \frac{\displaystyle\sum_{i=1}^{N}\sum_{j=1}^{M}
                   \bigl(X_{\mathrm{model}}(\mathbf{x}_{ij}) 
                   - \overline{X}_{\mathrm{model}}\bigr)^{2}}
                  {\displaystyle\sum_{i=1}^{N}\sum_{j=1}^{M}
                   \bigl(X_{\mathrm{obs}}(\mathbf{x}_{ij})   
                   - \overline{X}_{\mathrm{obs}}\bigr)^{2}},
  \label{eq:variance_ratio}
\]
where $\mathbf{x}_{ij}$ denotes the grid point at latitude index $i$ and longitude index $j$, with $i = 1, \dots, N$ and $j = 1, \dots, M$. $X_{\mathrm{model}}(\mathbf{x}_{i})$ and $X_{\mathrm{obs}}(\mathbf{x}_{i})$ denote the model-derived and observation-derived energy values at grid point $\mathbf{x}_{ij}$, and $\overline{X}_{\mathrm{model}}$, $\overline{X}_{\mathrm{obs}}$ are their respective
spatial means. A variance ratio closer to~1 indicates better agreement in spatial variability amplitude between the model and observations. Values smaller than~1 indicate that the modelled field is overly smooth, while values larger than~1 indicate excessive variability.\newline
\textbf{Amplitude ratio} \newline
We use the amplitude ratio to evaluate whether the mean intensity of the modeled field is consistent with the observations. It is formulated as follows:

\[
\mathrm{Amplitude\ Ratio}(X)
    = \frac{\overline{X}_{\mathrm{model}}}
           {\overline{X}_{\mathrm{obs}}}
    = \frac{
        \dfrac{1}{N\times M}\displaystyle\sum_{i=1}^{N}\sum_{j=1}^{M}
          X_{\mathrm{model}}(\mathbf{x}_{ij})
      }{
        \dfrac{1}{N\times M}\displaystyle\sum_{i=1}^{N}\sum_{j=1}^{M}
          X_{\mathrm{obs}}(\mathbf{x}_{ij})
      },
  \label{eq:amplitude_ratio}
\]
where $\overline{X}_{\mathrm{model}}$ and $\overline{X}_{\mathrm{obs}}$ denote the spatial mean values of the model-derived and observation-derived energy fields, respectively. An amplitude ratio closer to~1 indicates better agreement in the mean energy intensity between the model and observations.\newline
\textbf{Area Integral Ratio}\newline
We use the area integral ratio to evaluate the consistency of the 
total integrated energy between the modeled and observed fields. 
It is formulated as follows:

\[
    \mathrm{Area\ Integral\ Ratio}(X) = \frac{\displaystyle\int X_{\mathrm{model}}\, \mathrm{d}A}
                  {\displaystyle\int X_{\mathrm{obs}}\, \mathrm{d}A}
           = \frac{\displaystyle\sum_{i,j} A_{ij}\, 
                   X_{\mathrm{model}}(\mathbf{x}_{ij})}
                  {\displaystyle\sum_{i,j} A_{ij}\, 
                   X_{\mathrm{obs}}(\mathbf{x}_{ij})},
\]

where $A_{ij}$ is the spherical area of grid cell $(i,j)$, 
computed as

\[
    A_{ij} = R^{2}\,\Delta\lambda
             \left[
                 \sin\!\left(\phi_{i}+\frac{\Delta\phi}{2}\right)
                -\sin\!\left(\phi_{i}-\frac{\Delta\phi}{2}\right)
             \right],
\]

with $R$ being the Earth's radius, $\phi_i$ the latitude of grid point $i$, and $\Delta\phi$, $\Delta\lambda$ the grid spacings in 
latitude and longitude, respectively. An area integral ratio closer to 1 indicates better agreement between the modeled and observed fields. An area integral ratio closer to~1 indicates better agreement in the total integrated energy between the model and observations.\newline

\subsection*{Supplementary Note 4-Mesoscale Eddies Evaluation Metrics}
Based on the principles outlined in measures of forecast quality\citep{Metrics} for dichotomous (yes–no) forecasts, we employ Bias, Critical Success Index (CSI), False Alarm Ratio (FAR), and Probability of Detection (POD) to evaluate the forecast accuracy of mesoscale eddies.
To evaluate the forecast quality of mesoscale eddies, we define three categorical outcomes based on the correspondence between forecasts and observations:

A = number of eddies correctly forecast (hit)

B = number of eddies forecast but not observed (false alarm),

C = number of eddies observed but not forecast (miss),

From these, the following standard quality measures are defined:\newline
\textbf{Bias}\newline
$\mathrm{Bias}=\frac{A+B}{A+C}$ evaluates the systematic tendency of the forecast to overpredict or underpredict mesoscale eddy occurrence. A bias value of 1 represents an unbiased forecast, values greater than 1 indicate overforecasting, and values less than 1 indicate underforecasting.\newline
\textbf{Critical Success Index (CSI)}\newline
$\mathrm{CSI}=\frac{A}{A+B+C}$ provides an overall measure of forecast accuracy by jointly considering hits, misses, and false alarms while excluding correct negatives. A higher CSI indicates better overall forecast performance.\newline
\textbf{False Alarm Ratio (FAR)}\newline
$\mathrm{FAR}=\frac{B}{A+B}$ quantifies the proportion of forecasted mesoscale eddies that did not actually occur. A lower FAR indicates fewer false detections and therefore higher forecast reliability.\newline
\textbf{Probability of Detection (POD)}\newline
$\mathrm{POD} = \frac{A}{A+C}$ is the fraction of observed eddies that are successfully predicted and measures the ability of the forecasting system to correctly identify observed mesoscale eddies. A higher POD indicates that a larger proportion of actual eddies are successfully detected, reflecting the forecast's sensitivity to eddy occurrence.

\subsection*{Supplementary Note 5-Energy Calculation}
The total kinetic energy (KE) represents the overall intensity of ocean surface currents, including both the large-scale mean circulation and mesoscale eddy variability. Let $u_n(\mathbf{x})$ and $v_n(\mathbf{x})$ denote the zonal and meridional 
velocity components at grid point $\mathbf{x}$ on day $n$ 
($n = 1, 2, \ldots, N$). 
The time-mean kinetic energy per unit mass at each grid point is defined as:

\[
    \mathrm{KE}(\mathbf{x}) = \frac{1}{N} \sum_{n=1}^{N} 
    \frac{1}{2}\left[u_n^2(\mathbf{x}) + v_n^2(\mathbf{x})\right]
\label{eq:ke}
\]

The mean kinetic energy (MKE) represents the strength of the large-scale mean circulation. It is calculated from the temporally averaged velocity field, and the mean velocity components are
\[
\overline{u}(\mathbf{x})
=
\frac{1}{N}
\sum_{n=1}^{N}
u_n(\mathbf{x})
\]

\[
\overline{v}(\mathbf{x})
=
\frac{1}{N}
\sum_{n=1}^{N}
v_n(\mathbf{x})
\]
Then the mean kinetic energy (MKE) is defined as
\[
\mathrm{MKE}(\mathbf{x}) = \frac{1}{2}(\overline{u}^2(\mathbf{x}) + \overline{v}^2(\mathbf{x}))
\]

where \(\overline{u}\) and \(\overline{v}\) denote the temporal mean zonal and meridional velocity components, respectively.

The eddy kinetic energy (EKE) represents the intensity of mesoscale eddies and reflects mesoscale variability, flow instability, and energy cascade processes. It is defined as

\[
\mathrm{EKE}(\mathbf{x}) =
\frac{1}{N}\sum_{n=1}^{N}\frac{1}{2}
\left[
(u_n(\mathbf{x})-\overline{u}_n(\mathbf{x}))^2
+
(v_n(\mathbf{x})-\overline{v}_n(\mathbf{x}))^2
\right]
\]

where \(u-\overline{u}\) and \(v-\overline{v}\) represent the velocity anomalies relative to the temporal mean flow.

\newpage

\begin{figure}[H]
    \centering
    \includegraphics[width=\textwidth]{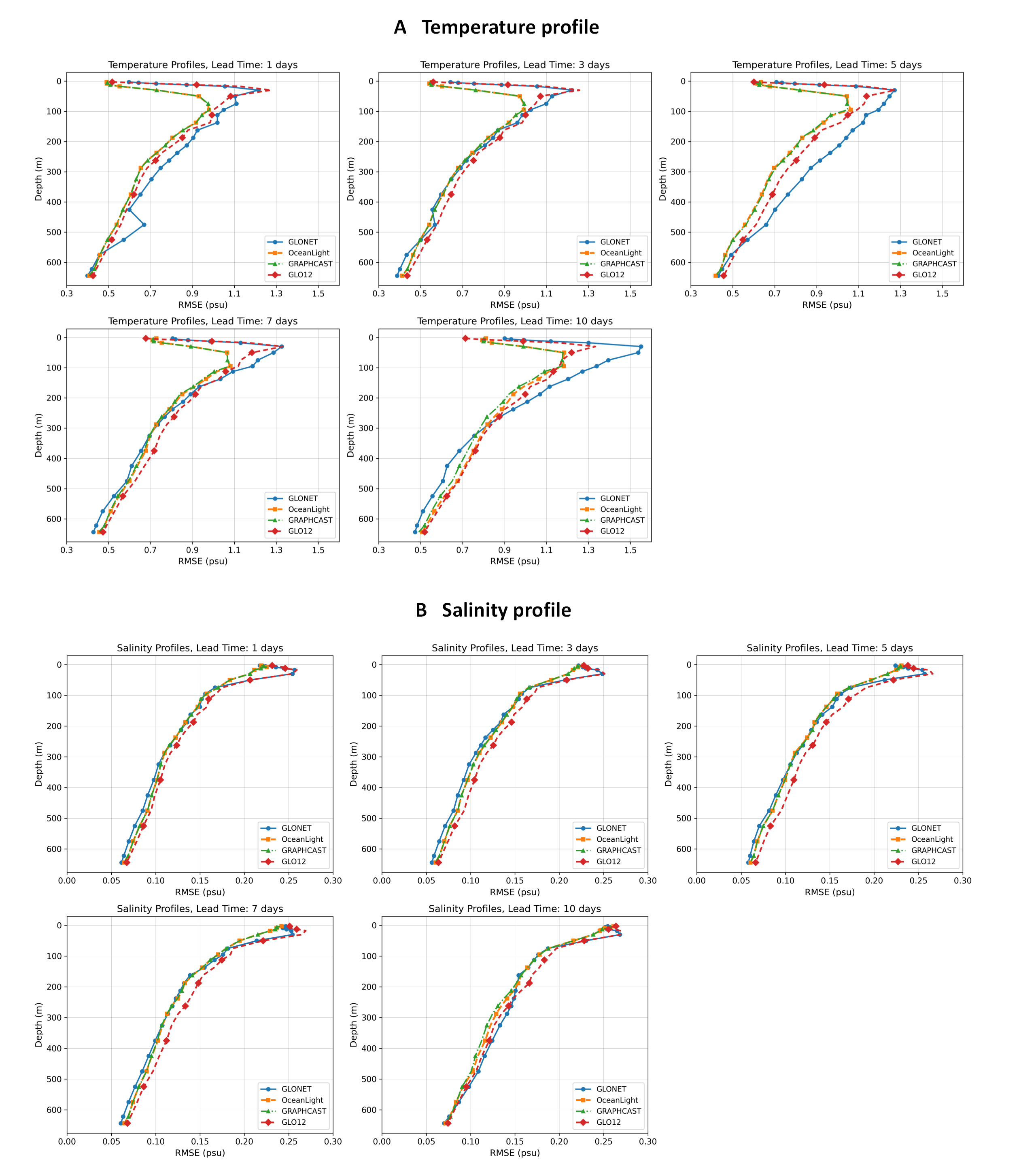}
    \caption{Vertical RMSE profiles of (A) temperature and (B) salinity as a function of depth for lead times of 1, 3, 5, 7, and 10 days, comparing GLONET, OceanLight(our model), GraphCast, and the GLO12 baseline. For each variable, RMSE generally increases near the surface and in the upper thermocline/halocline (0–150 m) before decreasing with depth, with the magnitude of near-surface errors growing progressively with longer lead times. OceanLight and GraphCast track each other closely across all depths and lead times, consistently achieving lower RMSE than GLONET, particularly in the subsurface temperature maximum region (50–300 m), while remaining comparable to or better than the GLO12 baseline. Salinity RMSE profiles show a similar depth-dependent pattern across models, with smaller inter-model differences than for temperature.}
    \label{fig:tempsalprofile}
\end{figure}

\begin{figure}[H]
    \centering
    \includegraphics[width=\textwidth]{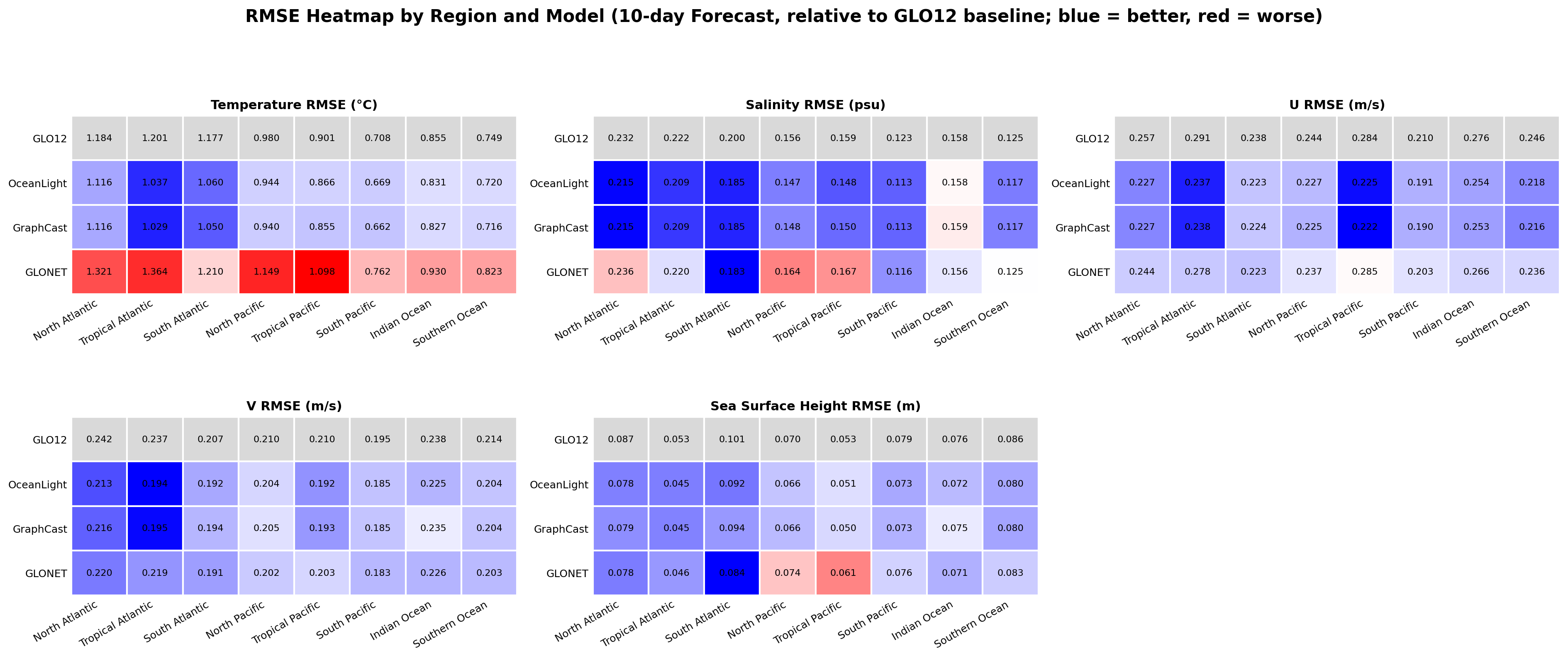}
    \caption{Regional RMSE comparison for 10-day forecasts across five variables — temperature, salinity, zonal velocity (U), meridional velocity (V), and sea surface height — evaluated over eight ocean basins. Values are shown for the GLO12 numerical baseline (grey, top row) and three data-driven models: OceanLight (ours), GraphCast, and GLONET. Cell colors indicate RMSE relative to the GLO12 baseline within each variable panel, with blue denoting improvement and red denoting degradation. OceanLight achieves consistent RMSE reductions relative to GLO12 across nearly all regions and variables, with performance closely matching GraphCast and generally surpassing GLONET.}
    \label{fig:heatmap}
\end{figure}

\begin{figure}[H]
    \centering
    \includegraphics[width=\textwidth]{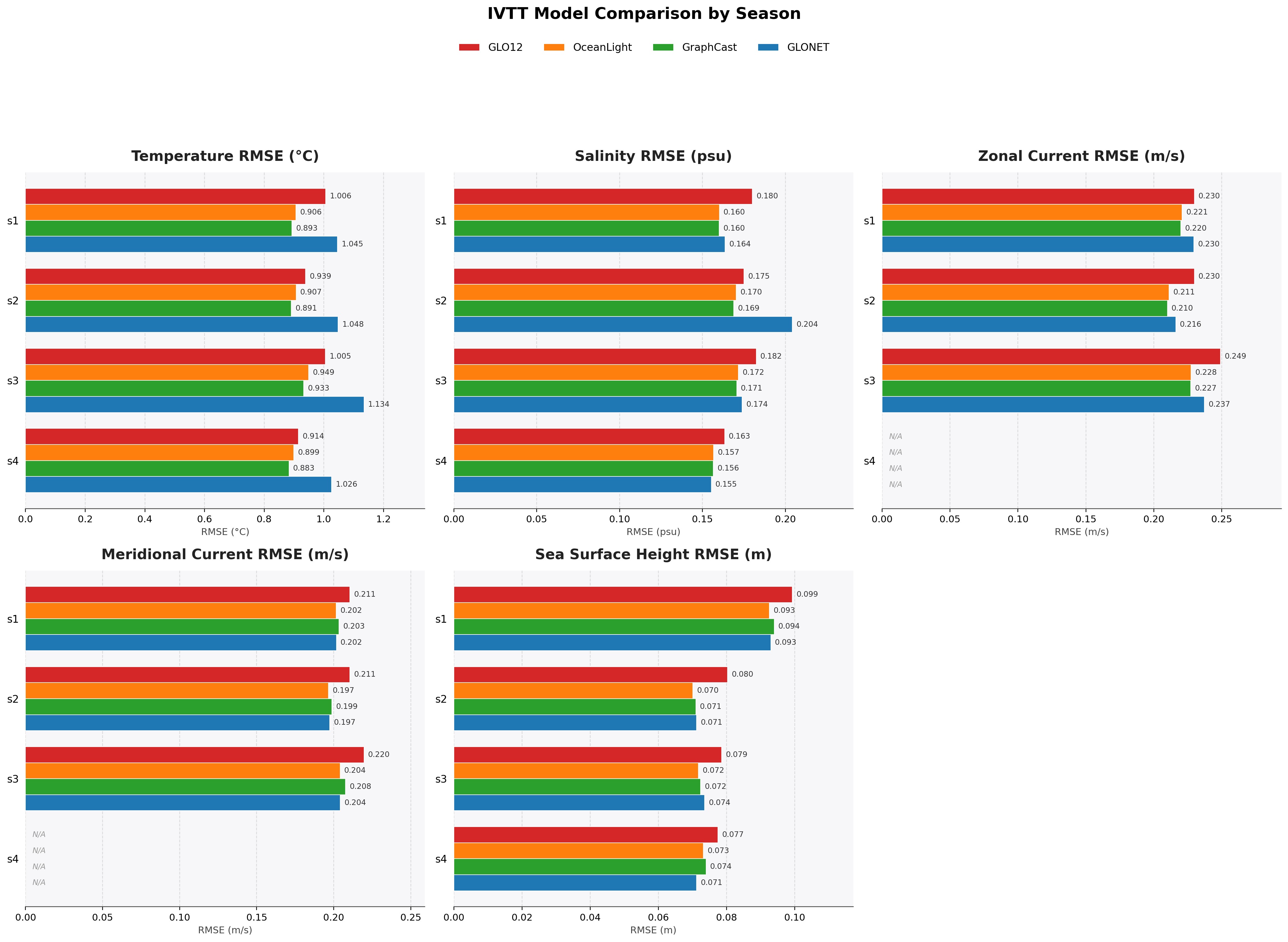}
    \caption{RMSE comparison of GLO12, OceanLight, GraphCast, and GLONET across four seasonal evaluation periods (s1–s4), for temperature, salinity, zonal/meridional current velocity, and sea surface height. Current velocity metrics are unavailable for s4 (N/A) due to missing observational data for that period.}
    \label{fig:seasonbat}
\end{figure}

\begin{figure}[H]
    \centering
    \caption*{\footnotesize\textbf{Kuroshio and its Extension Region}}
    \includegraphics[width=\textwidth]{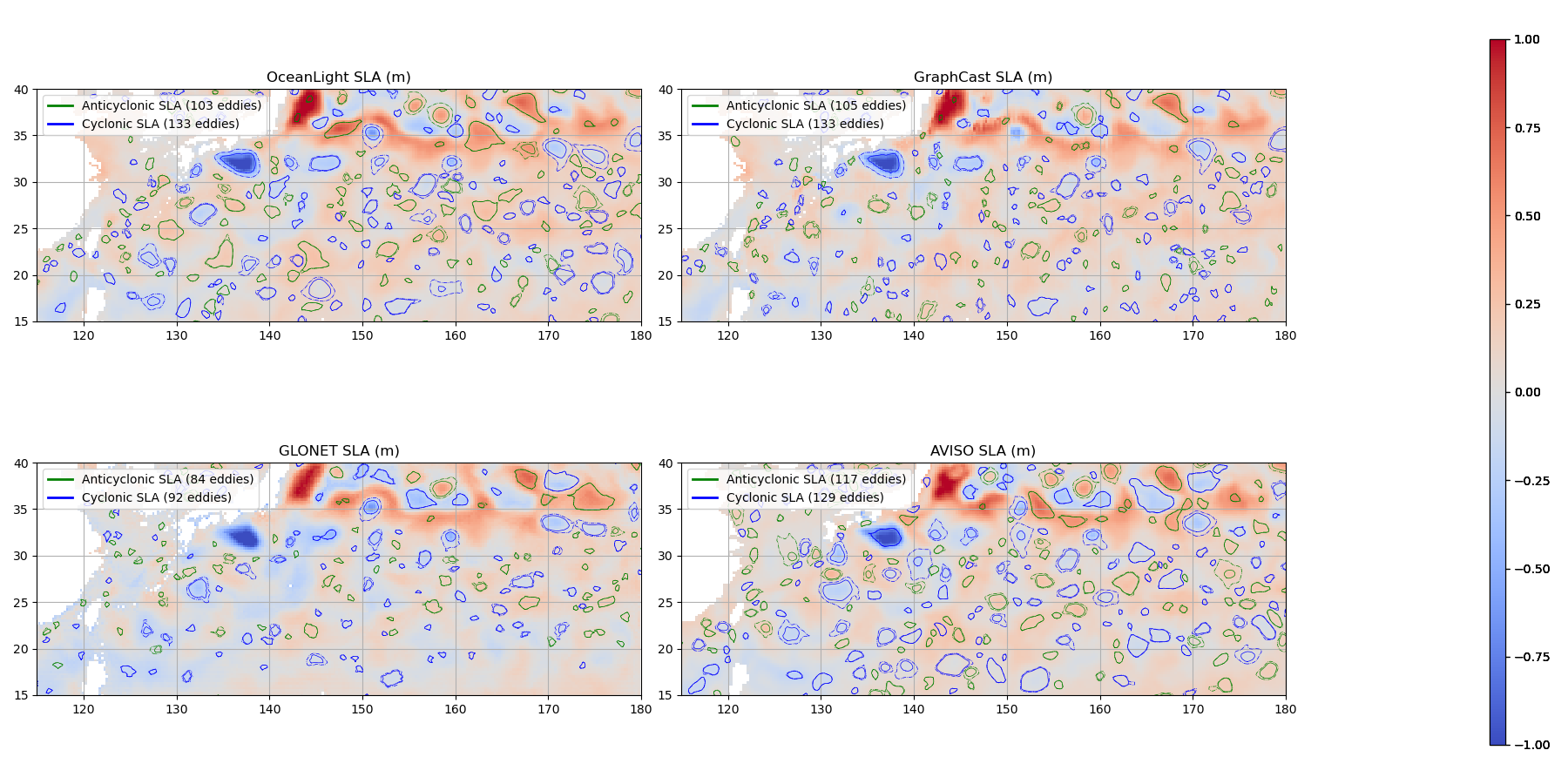}
    \caption{Mesoscale eddy visualization in the Kuroshio and its extension region. Forecasts are initialized from OceanBench of January 2nd, 2024 with a 10-day lead time, and validated against AVISO observations of January 12th, 2024. Eddy structures are detected using the PET algorithm applied consistently to both forecasted and observed SLA fields.}
    \label{fig:eddy1}
\end{figure}

\begin{figure}[H]
    \centering
    \caption*{\footnotesize\textbf{Gulf Stream}}
    \includegraphics[width=\textwidth]{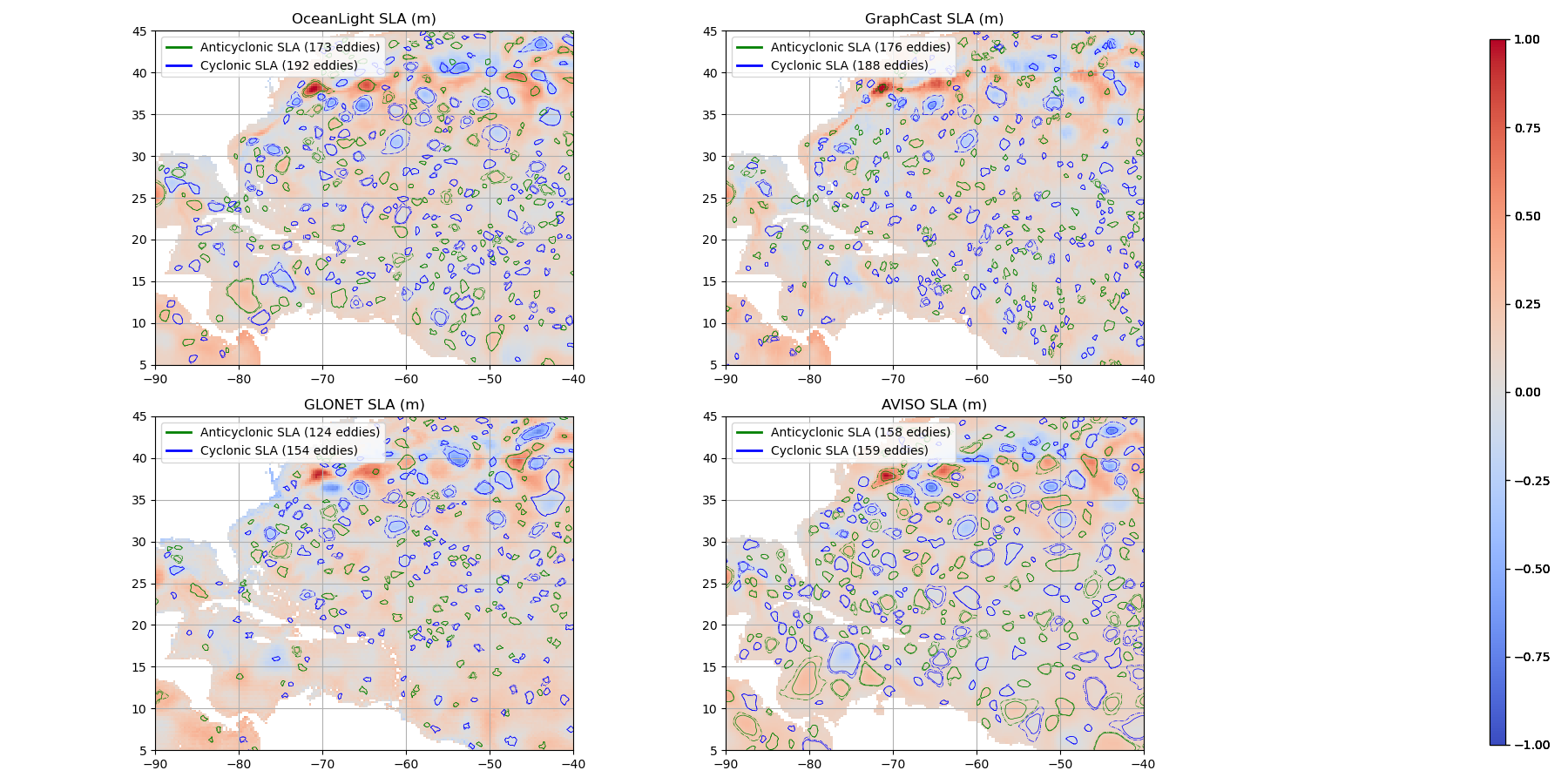}
    \caption{Mesoscale eddy visualization in the Gulf Stream. Forecasts are initialized from OceanBench of January 2nd, 2024 with a 10-day lead time, and validated against AVISO observations of January 12th, 2024. Eddy structures are detected using the PET algorithm applied consistently to both forecasted and observed SLA fields.}
    \label{fig:eddy2}
\end{figure}

\begin{figure}[H]
    \centering
    \caption*{\footnotesize\textbf{Brazil Current}}
    \includegraphics[width=\textwidth]{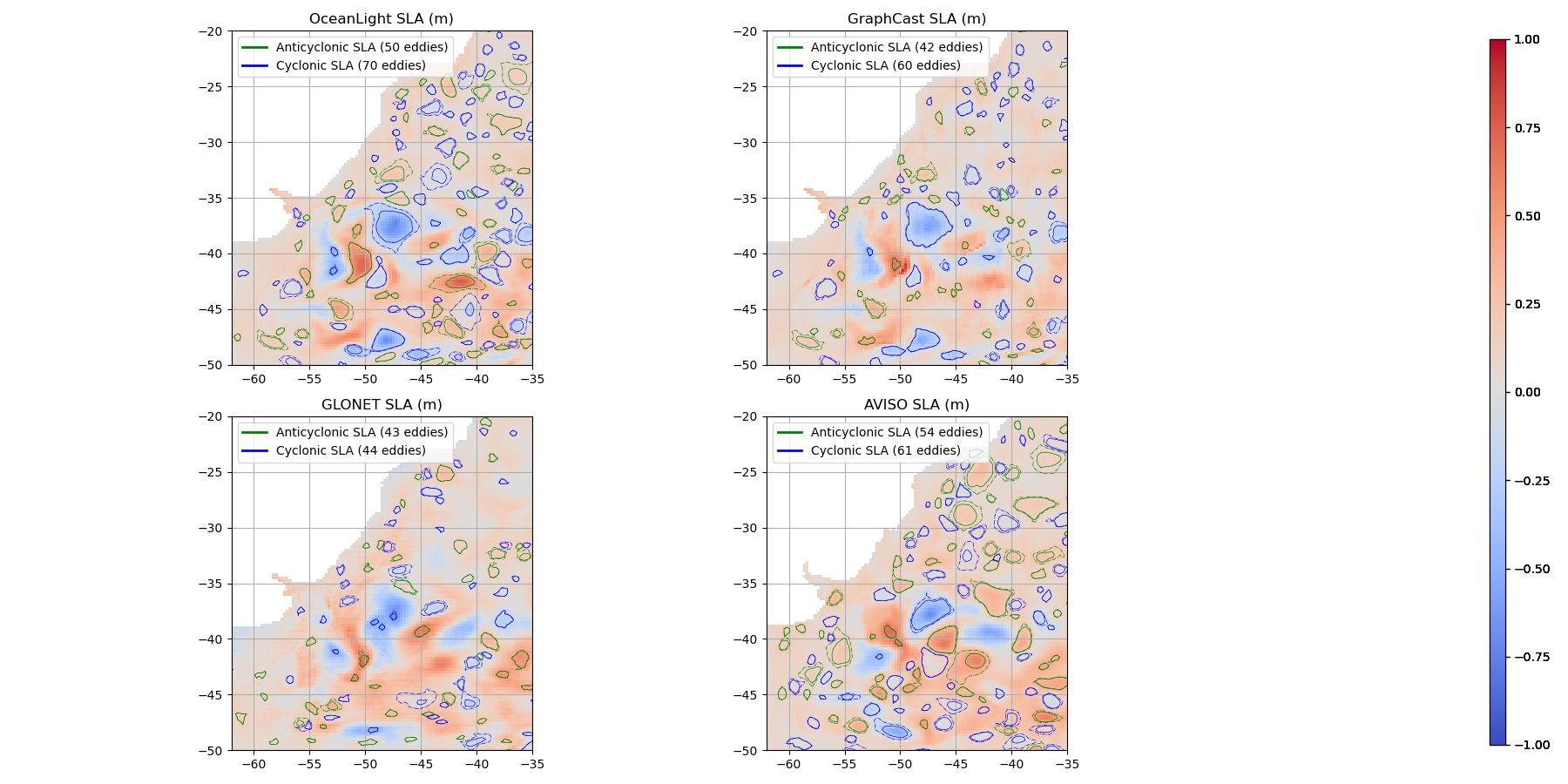}
    \caption{Mesoscale eddy visualization in the Brazil Current. Forecasts are initialized from OceanBench of January 2nd, 2024 with a 10-day lead time, and validated against AVISO observations of January 12th, 2024. Eddy structures are detected using the PET algorithm applied consistently to both forecasted and observed SLA fields.}
    \label{fig:eddy3}
\end{figure}

\begin{figure}[H]
    \centering   \includegraphics[width=\textwidth]{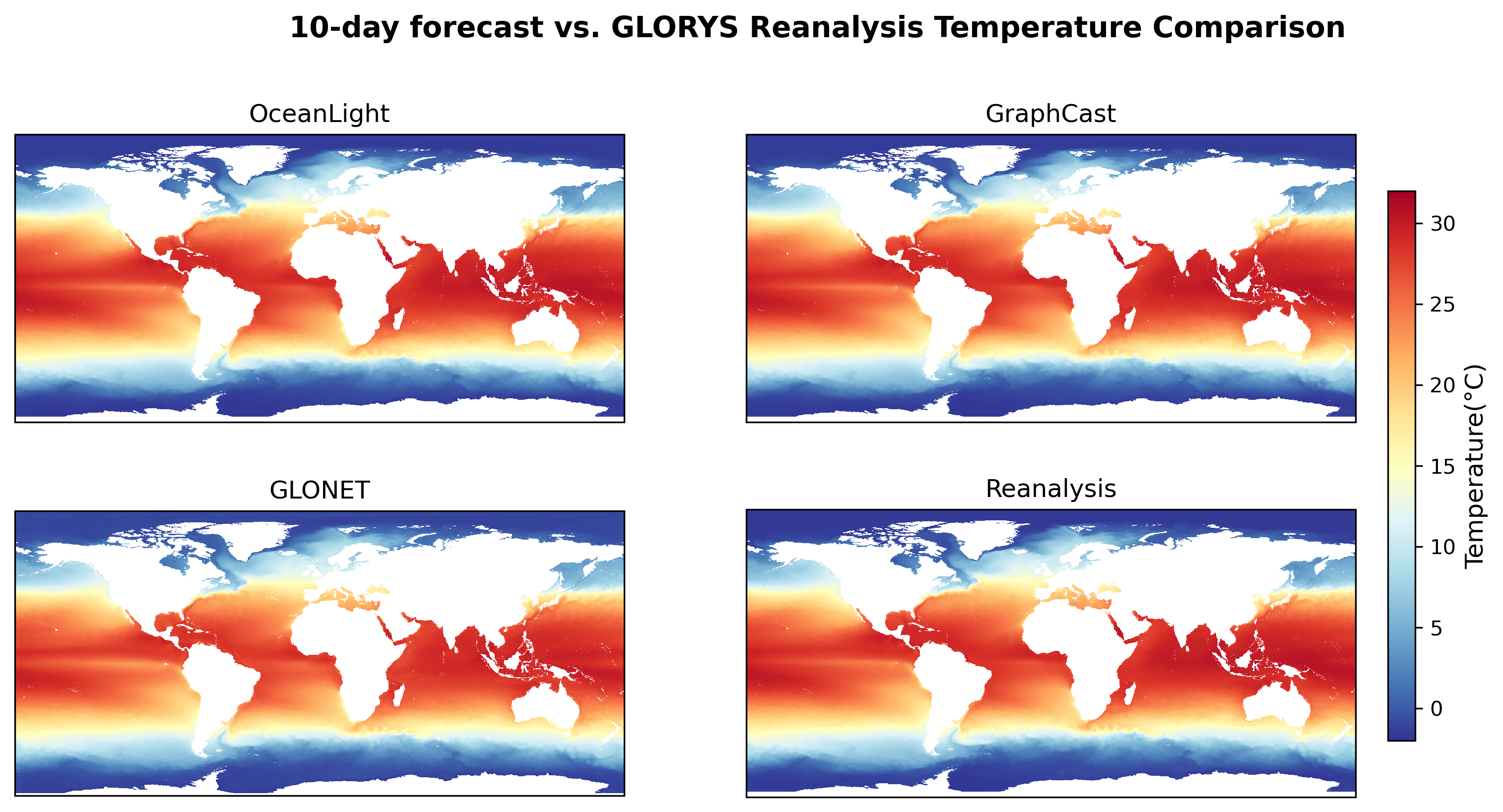}
    \caption{Global sea surface temperature fields averaged over 10-day forecasts for all of 2024, using OceanBench-derived initial conditions, for OceanLight (ours), GraphCast, and GLONET, compared against the corresponding time-averaged GLORYS reanalysis field.}
    \label{fig:thetaovisual}
\end{figure}

\begin{figure}[H]
    \centering   \includegraphics[width=\textwidth]{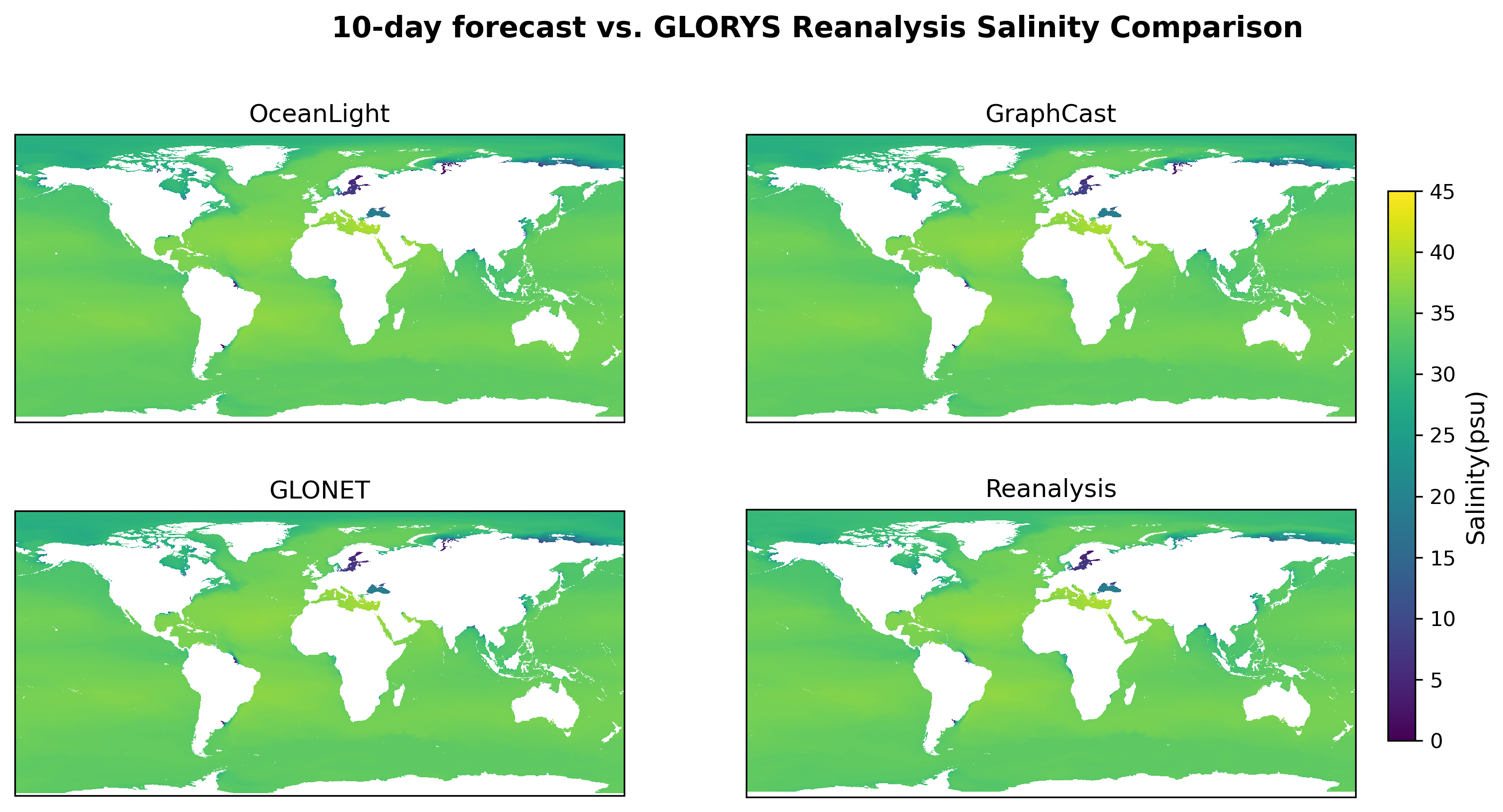}
    \caption{Global sea surface salinity fields averaged over 10-day forecasts for all of 2024, using OceanBench-derived initial conditions, for OceanLight (ours), GraphCast, and GLONET, compared against the corresponding time-averaged GLORYS reanalysis field.}
    \label{fig:sovisual}
\end{figure}

\begin{figure}[H]
    \centering   \includegraphics[width=\textwidth]{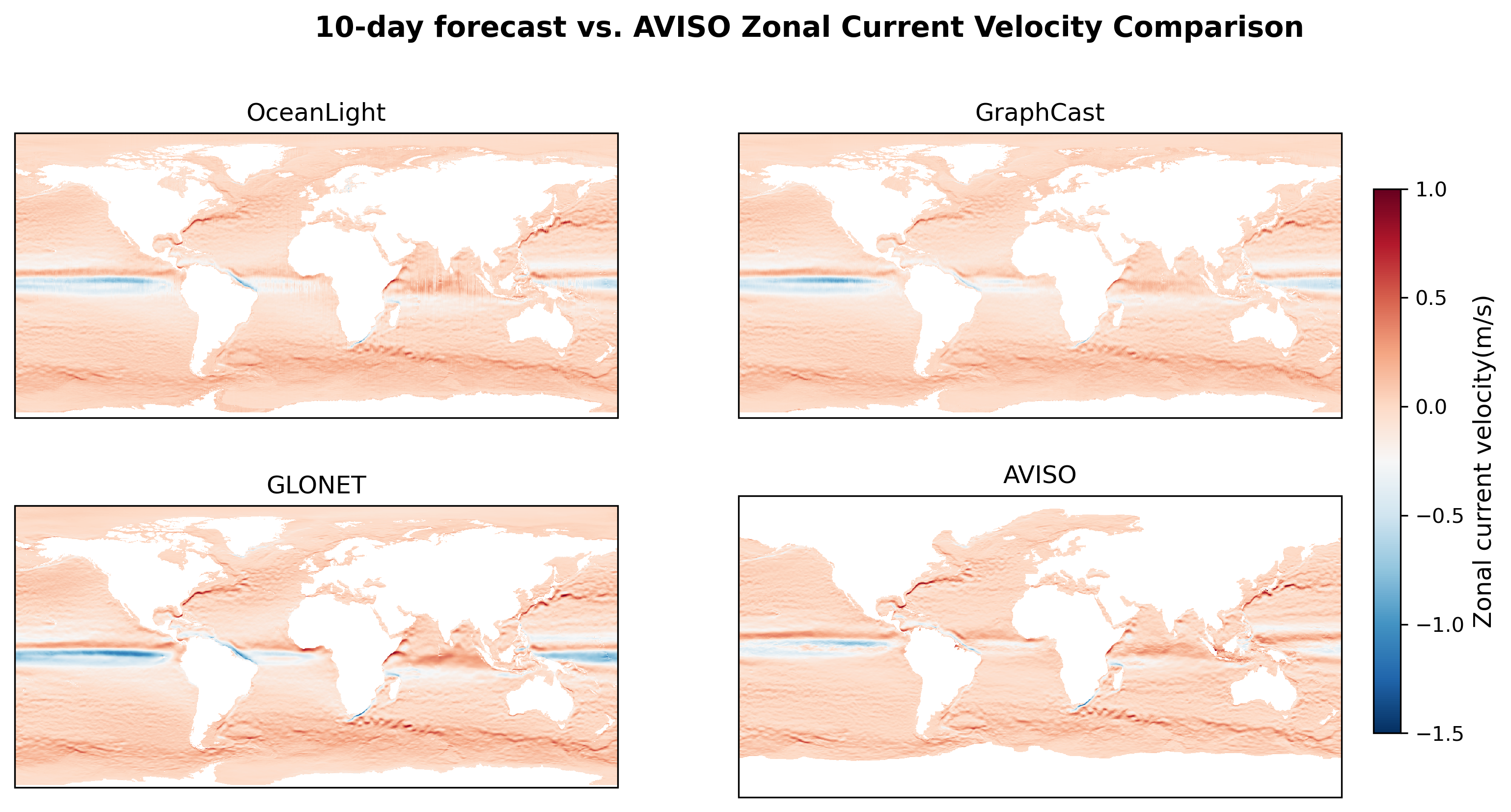}
    \caption{Global sea surface zonal current velocity fields averaged over 10-day forecasts for all of 2024, using OceanBench-derived initial conditions, for OceanLight (ours), GraphCast, and GLONET, compared against the corresponding time-averaged GLORYS reanalysis field.}
    \label{fig:uovisual}
\end{figure}

\begin{figure}[H]
    \centering   \includegraphics[width=\textwidth]{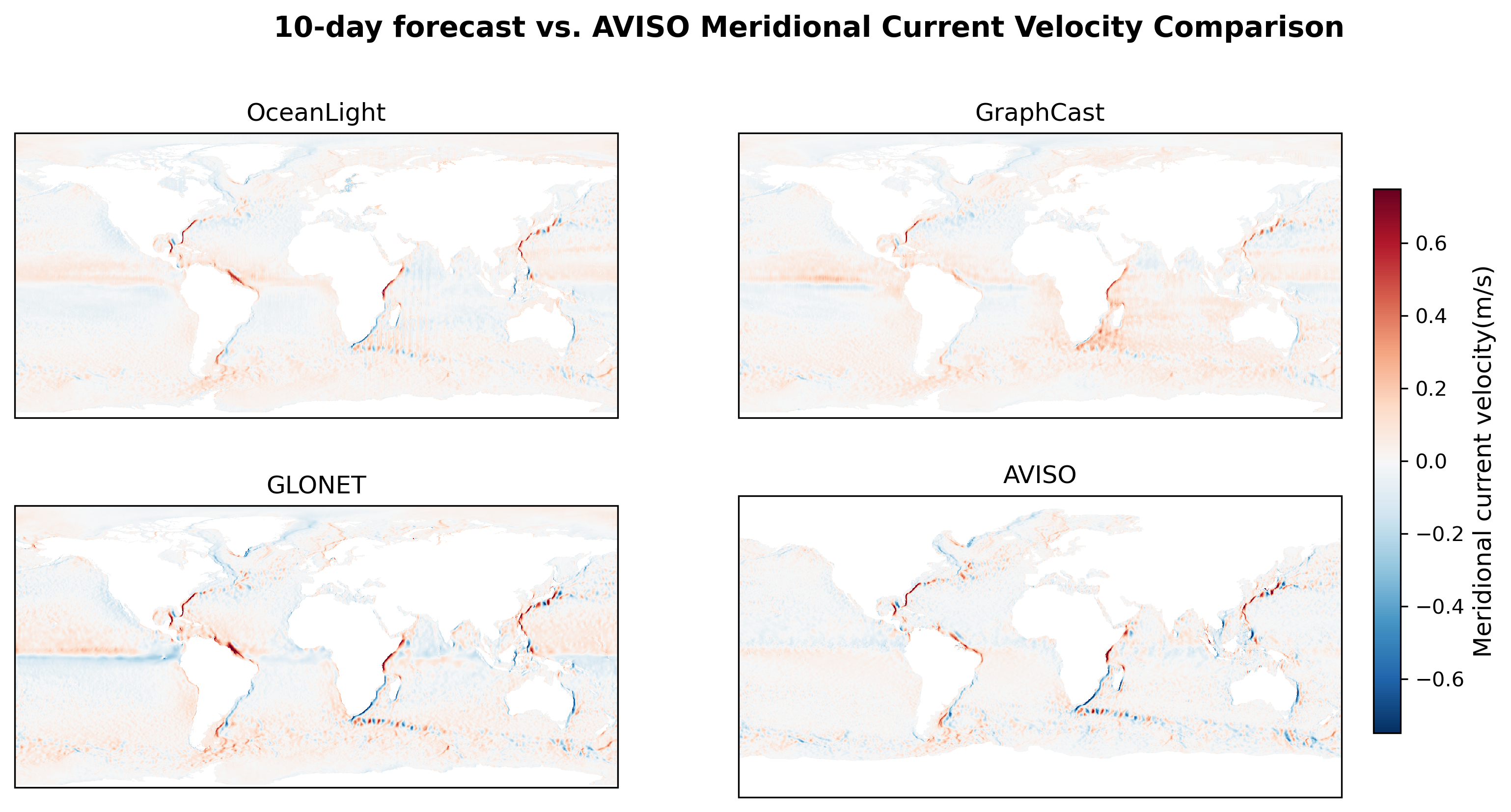}
    \caption{Global sea surface meridional current velocity fields averaged over 10-day forecasts for all of 2024, using OceanBench-derived initial conditions, for OceanLight (ours), GraphCast, and GLONET, compared against the corresponding time-averaged GLORYS reanalysis field.}
    \label{fig:vovisual}
\end{figure}

\begin{figure}[H]
    \centering   \includegraphics[width=\textwidth]{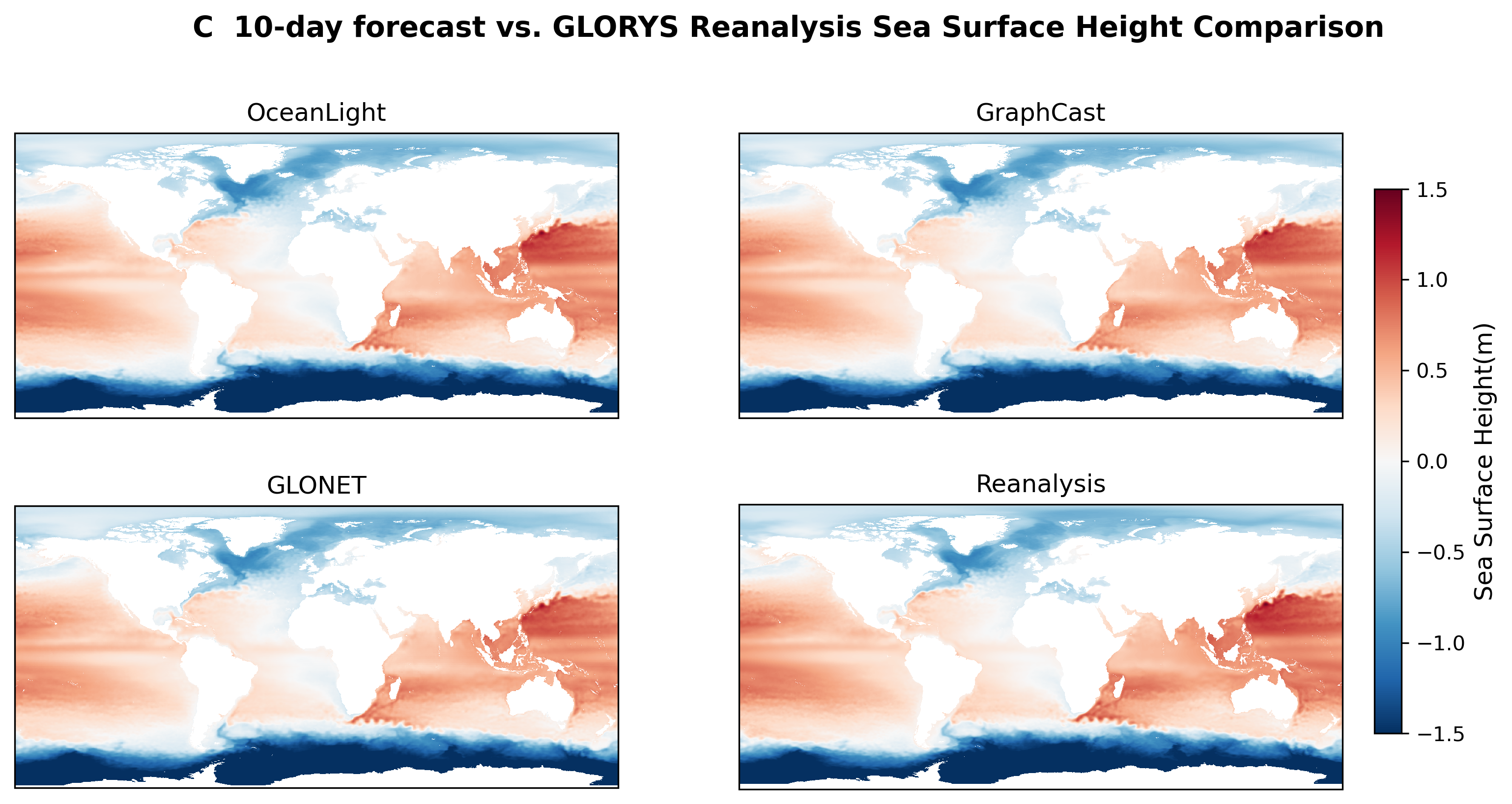}
    \caption{Global sea surface height fields averaged over 10-day forecasts for all of 2024, using OceanBench-derived initial conditions, for OceanLight (ours), GraphCast, and GLONET, compared against the corresponding time-averaged GLORYS reanalysis field.}
    \label{fig:zosvisual}
\end{figure}

\begin{figure}[H]
    \centering   \includegraphics[width=\textwidth]{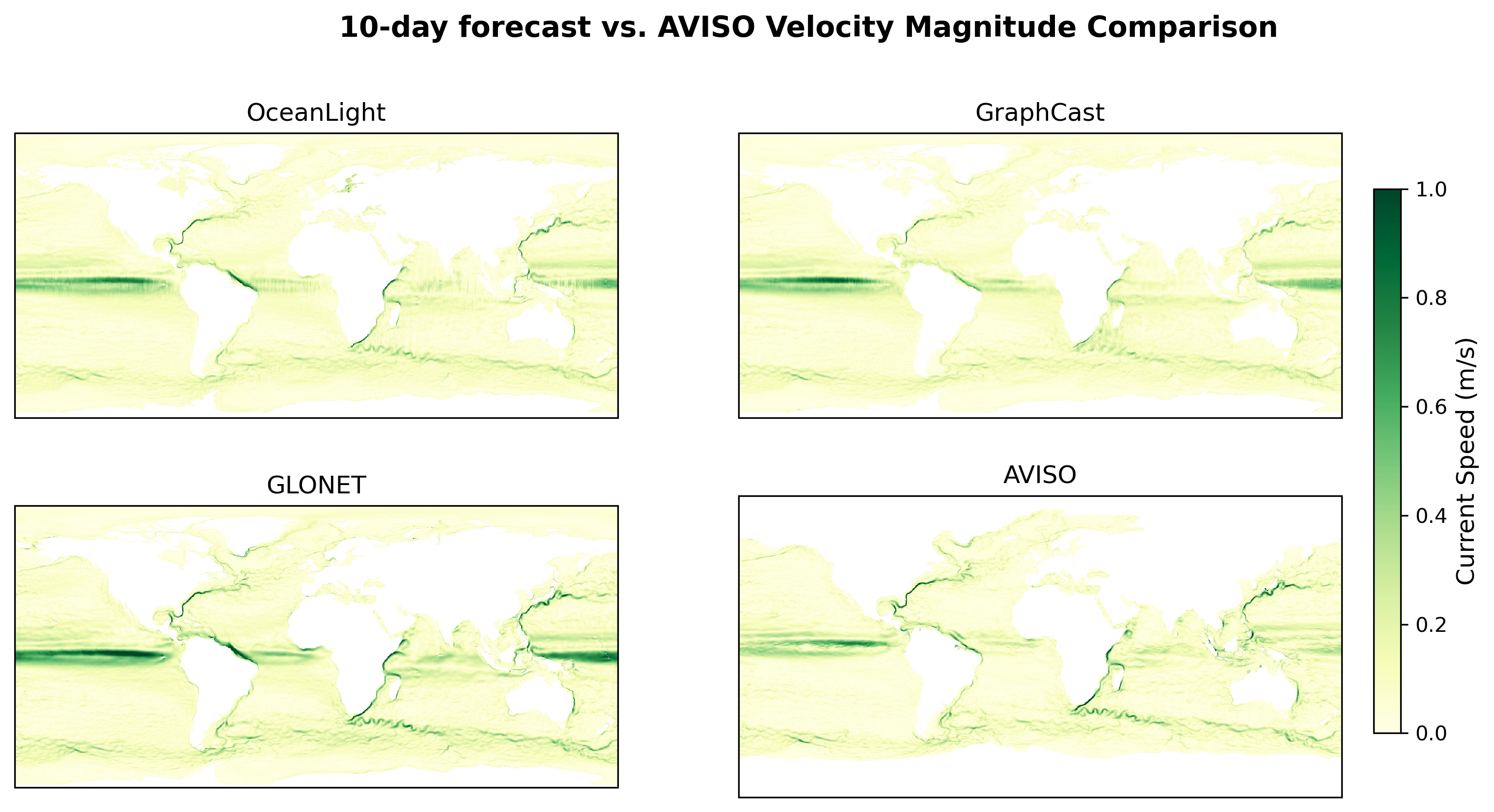}
    \caption{Comparison of 10-day forecast sea surface current speed (magnitude of the horizontal velocity, $|\mathbf{u}| = \sqrt{u^2+v^2}$) from three models---OceanLight, GraphCast, and GLONET---against AVISO satellite altimetry-derived observations.}
    \label{fig:velocitymagnitude}
\end{figure}

\begin{figure}[H]
    \centering   \includegraphics[width=0.8\textwidth]{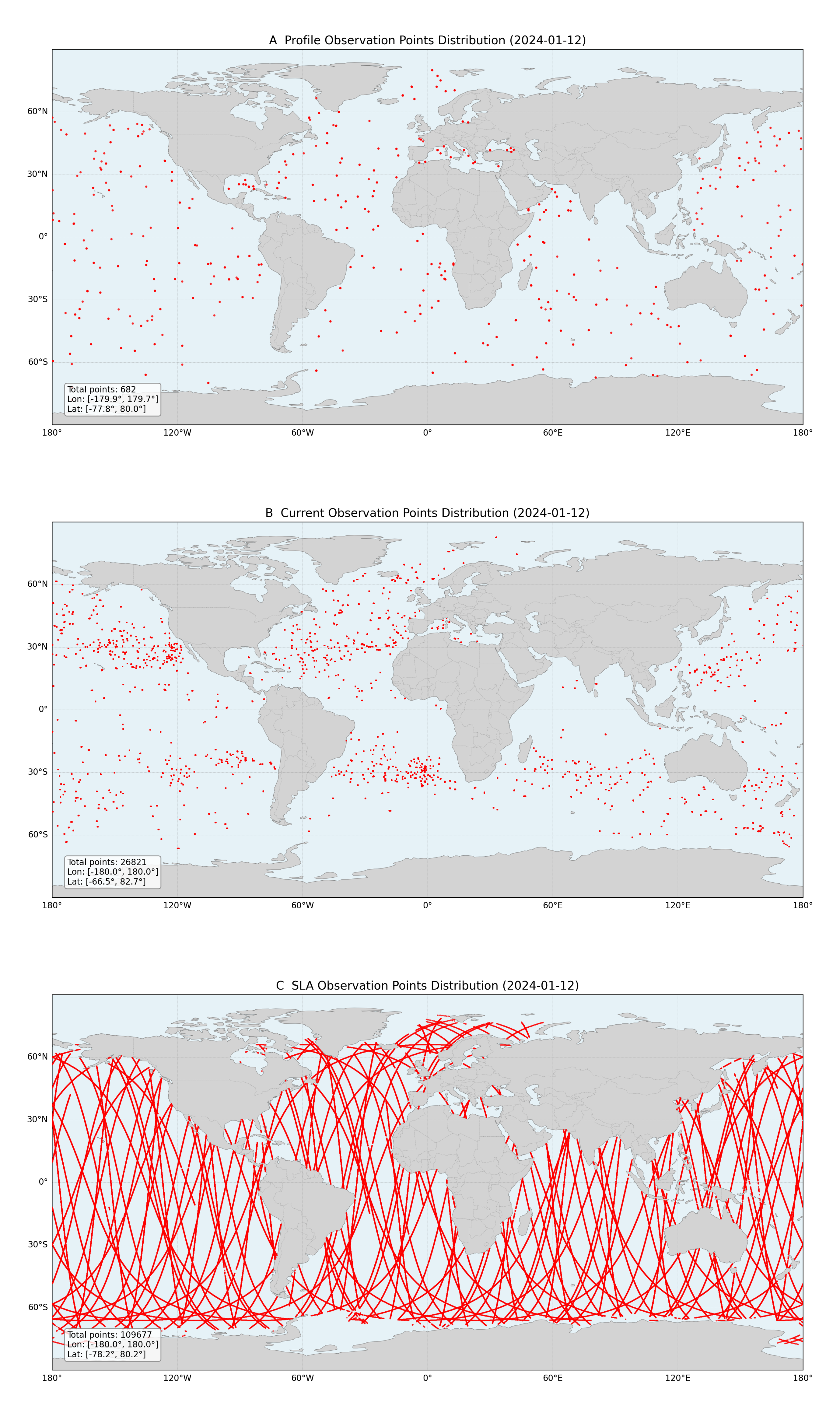}
    \caption{Spatial distribution of IV-TT Class4 reference observations used for global ocean forecast verification, obtained from the OceanPredict (formerly GODAE OceanView) Intercomparison and Validation Task Team (IV-TT) archive hosted on the NCI THREDDS server, for 2024-01-12. (A) Sub-surface profile observations of 682 points, sparsely distributed across the global ocean. (B) In situ current (velocity) observations of 26,821 points, showing denser sampling concentrated along subtropical gyres and western boundary current regions. (C) Along-track satellite altimeter sea level anomaly (SLA) observations of 109,677 points, exhibiting the characteristic crosshatch pattern of overlapping ascending/descending orbital passes with near-global coverage.}
    \label{fig:ivttvisual}
\end{figure}

\begin{figure}[H]
    \centering
    \includegraphics[width=\linewidth]{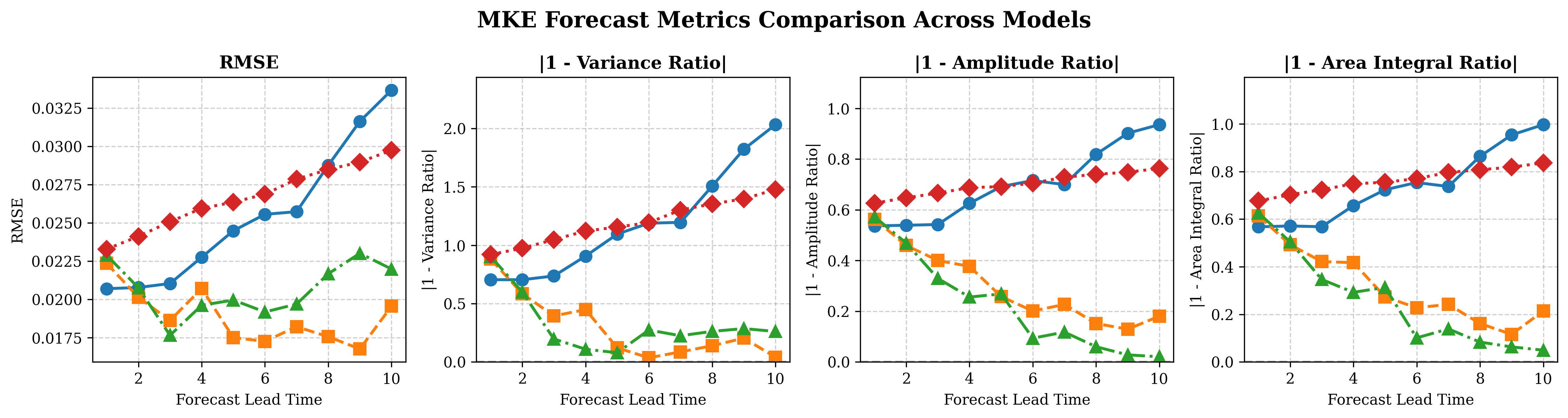}
    \caption{Mean kinetic energy (MKE) are computed from the forecasted surface velocity fields and compared against the corresponding energy fields derived from AVISO altimetry, as a function of forecast lead time. For each energy component, four metrics are evaluated: root-mean-square error (RMSE), $|1-\text{Variance Ratio}|$, $|1-\text{Amplitude Ratio}|$, and $|1-\text{Area Integral Ratio}|$, where the three ratio-based metrics are expressed as absolute deviations from unity such that lower values indicate closer agreement with the AVISO-derived reference fields.}
    \label{fig:mke}
\end{figure}

\begin{figure}[H]
    \centering
    \includegraphics[width=\linewidth]{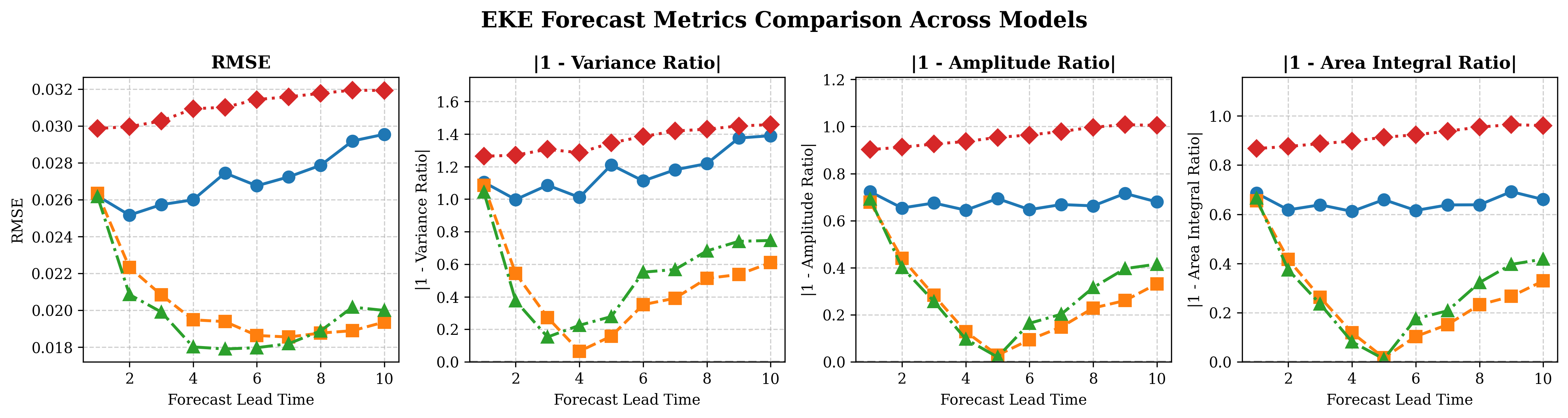}
    \caption{Eddy kinetic energy (EKE) are computed from the forecasted surface velocity fields and compared against the corresponding energy fields derived from AVISO altimetry, as a function of forecast lead time.}
    \label{fig:mke}
\end{figure}


\end{document}